\documentclass[11pt]{article}
\usepackage[a4paper,total={6.5in,9.5in}]{geometry}
\usepackage[english]{babel}

\usepackage[utf8]{inputenc} 
\usepackage[T1]{fontenc}    
\usepackage[colorlinks=true,linkcolor=blue]{hyperref}      
\usepackage{url}           
\usepackage{booktabs}       
\usepackage{amsfonts}      
\usepackage{nicefrac}      
\usepackage{microtype}
\usepackage[dvipsnames]{xcolor}

\usepackage{float}
\usepackage{graphics}
\usepackage{subcaption}
\graphicspath{{./figures/}}
\DeclareGraphicsExtensions{.pdf}

\usepackage{algorithm}
\usepackage{algpseudocode}

\usepackage{mathtools}
\usepackage{amsmath}
\usepackage{bm}

\usepackage{amsthm}
\usepackage{amssymb}
\usepackage[capitalize]{cleveref}

\usepackage[numbers]{natbib}

\usepackage{nicefrac}

\usepackage{comment}

\usepackage{amsthm}
\usepackage{graphicx}

\usepackage{float} 
\usepackage{indentfirst}
\usepackage{appendix}
\usepackage{float}

\newtheorem{assumption}{Assumption}
\newtheorem{assumptionalt}{Assumption}[assumption]

\newenvironment{assumptionp}[1]{
  \renewcommand\theassumptionalt{#1}
  \assumptionalt
}{\endassumptionalt}
\newtheorem{example}{Example}[section]
\newtheorem{theorem}{Theorem}
\theoremstyle{remark}
\newtheorem{remark}{Remark}
\newcommand{\ci}[3]{#1 \perp\kern-5pt \perp #2 \mid #3}
\theoremstyle{definition}
\newtheorem{definition}{Definition}[section]

\title{Comparing Model-agnostic Feature Selection Methods through Relative Efficiency}

\author{Chenghui Zheng
\thanks{Department of Statistics, University of Wisconsin - Madison, Madison, WI 53706}
\and 
Garvesh Raskutti
\footnotemark[1]
}

\date{\vspace{-5ex}}

\begin{document}
\maketitle

\begin{quotation}
\noindent {\it Abstract:}
Feature selection and importance estimation in a model-agnostic setting is an ongoing challenge of significant interest. Wrapper methods are commonly used because they are typically model-agnostic. In this paper, we develop a general comparison framework for model-agnostic feature selection methods based on relative efficiency, using \emph{relative variability} $\sigma/\mu$ to account for different statistics having different means. In particular we focus on state-of-the-art feature selection methods, the Generalized Covariance Measure (GCM) and Leave-One-Covariate-Out (LOCO) estimation.
In particular, we present a theoretical comparison under three model settings: linear models, non-linear additive models, and single index models that mimic a single-layer neural network. We complement this with simulations and real data examples for the above models and mis-specified models. Our theoretical results, along with empirical findings, demonstrate that GCM-related methods generally out-perform LOCO under suitable regularity conditions defined by a suitably defined correlation quantity which quantifies the asymptotic relative efficiency of these approaches.
Our simulations and real data analysis include widely used machine learning methods such as neural networks and gradient boosting trees.

\vspace{9pt}
\par
\end{quotation}\par

\def\thefigure{\arabic{figure}}
\def\thetable{\arabic{table}}

\renewcommand{\theequation}{\thesection.\arabic{equation}}

\fontsize{12}{14pt plus.8pt minus .6pt}\selectfont

\section{Introduction}
\label{sec:introduction}

Feature selection is one of the fundamental challenges in statistics and machine learning. While embedded methods such as Lasso have been widely used under parametric assumptions, there is a growing trend toward using “black-box” prediction methods such as random forests, neural networks, and others. In such settings, wrapper methods for feature selection which are typically \emph{model-agnostic} have become increasingly popular.

A natural question to consider is how we provide a theoretical comparison for model-agnostic feature selection methods. In this paper, we use relative efficiency based on the \emph{relative variability} $\sigma/\mu$ statistic to account for the fact that different estimators have different means. In particular we focus on two state-of-the-art wrapper methods: Leave-One-Covariate-Out (LOCO)~\cite{lei_distribution-free_2018} and the Generalized Covariance Measure (GCM)~\cite{shah_hardness_2020}. LOCO may be viewed as a generalization of ANOVA-based methods that measure the change in prediction error when a feature is removed~\cite{fisher_statistical_1992, duda_pattern_2000}. GCM, on the other hand, generalizes traditional correlation-based tests~\cite{spearman_proof_1904,pearson_vii_1997, szekely_measuring_2007} to conditional correlation, and can capture general functional relationships. While there has been prior work analyzing each of these methods and establishing asymptotic normality~\cite{shah_hardness_2020,williamson_general_2022}, there is no quantitative comparison of their performance in feature selection that we are aware of.

Concretely, consider a prediction model:
\[
E[Y \mid X_1, X_2, \ldots, X_p] = f(X_1, X_2, \ldots, X_p)
\]
for some unknown function $f$, where $Y$ is the response and $(X_1, X_2, \ldots, X_p)$ are the covariates. The goal of feature selection is to identify which covariates significantly affect $Y$. This enhances model interpretability, reduces overfitting, and can improve predictive performance.

In this paper, we compare GCM and LOCO with respect to statistical inference and practical applications under three statistical models: linear models, non-linear additive models and single-index models. We also include other model-agnostic variable importance methods such as:
the \emph{dropout method}, which replaces a feature with its marginal mean to avoid retraining;
\emph{Lazy-VI}~\cite{gao_lazy_2022}, which approximates the reduced model for efficiency while maintaining inferential guarantees.

The remainder of the paper is organized as follows. Section~\ref{VI_test} introduces variable importance tests and asymptotic relative efficiency test for the comparison between feature selection methods. Section~\ref{sec:models} compares LOCO, and GCM under linear, additive, and single index models. Section~\ref{sec:lazy} extends the comparison to dropout and Lazy-VI. Sections~\ref{sec:simulation} and~\ref{sec:real_data} present simulation studies and real data experiments, respectively. We conclude in Section~\ref{sec:conclusion}.

\subsection{Related work}
Feature selection techniques fall into three main categories: filter-based, wrapper-based, and embedded methods. \emph{Filter methods} involve calculating a relevance score for each feature independently of any model and rank features accordingly. They are computationally efficient but due to their simplicity often don't capture conditional relationships. Examples include mutual information~\cite{hoque_mifs-nd_2014}, information gain, Pearson correlation~\cite{pearson_vii_1997, yu_feature_2003}, the Chi-squared test~\cite{witten_practical_2005}, and the Fisher score~\cite{duda_pattern_2000}.\emph{Wrapper methods} involve evaluating subsets of features by training models and measuring performance, often using criteria like AIC, BIC, or Cp. Common algorithms include forward selection, backward elimination~\cite{kittler_feature_1978}, and recursive feature elimination~\cite{guyon_gene_2002}. While they account for feature interactions, they are computationally expensive and prone to overfitting due to repeated training. \emph{Embedded methods} perform feature selection during model training. Examples include Lasso~\cite{ma_penalized_2008}, elastic net~\cite{zou_regularization_2005}, decision trees, random forests~\cite{sandri_variable_2006}, and naive Bayes~\cite{cortizo_multi_2006}. Embedded methods are generally more efficient than wrappers but are tied to specific model classes.

Wrapper-style, model-agnostic feature selection has grown in popularity due to their flexibility and the increased use of so-called black-box methods. Among global model-agnostic approaches, permutation importance~\cite{breiman_random_2001} is the most widely used. It disrupts a feature's relationship with the response by permuting values and measuring the effect on predictive accuracy~\cite{strobl_conditional_2008, fisher_all_2019, mentch_quantifying_2016, altmann_permutation_2010}. However, these methods introduce randomness, which can lead to instability. The knockoff framework~\cite{barber_controlling_2015, candes_panning_2018} uses conditional randomization to control false discovery by generating synthetic ``null'' copies of features. While powerful, this approach relies on the assumption of feature exchangeability, which is difficult to verify in practice. Furthermore, the theoretical results focus on false discovery rate (FDR) rather than the mean and variance which is the focus of this paper. Replacing rather than permuting a feature, e.g., the dropout method, offers computational efficiency but may underfit~\cite{chang_dropout_2018}. LOCO avoids this issue by removing a feature and retraining, and has been generalized for conformal inference~\cite{lei_distribution-free_2018} and bootstrapping in high-dimensional settings~\cite{rinaldo_bootstrapping_2019}.

\paragraph{Conditional Independence Tests}
Feature selection directly relates to \emph{conditional independence testing}. In particular, any feature $X_j$ is informative to $Y$ if and only if:
$$
Y \not \perp X_j|X_1, X_2,...,X_{j-1}, X_{j+1}, X_{j+2},...,X_p
$$
or $Y$ is not independent of $X_j$ given all other variables/features. Classical tools like partial correlations rely on strong parametric assumptions. More flexible tests include Conditional randomization~\cite{candes_panning_2018}, Holdout randomization~\cite{tansey_holdout_2022}, Kernel-based~\cite{zhang_kernel-based_2012} and Mutual information-based~\cite{fleuret_fast_2004}. However, many of these methods are computationally expensive and rely on estimating conditional distributions. Generalized Covariance Measure (GCM)~\cite{shah_hardness_2020}, which tests for vanishing normalized covariance between residuals, is a regression-based test that requires only mild assumptions and works well in high dimensions and nonlinear settings.

This paper focuses on analyzing wrapper-based feature selection using LOCO and GCM. All methods yield population-level variable importance scores and support hypothesis testing. We provide theoretical and empirical comparisons across various settings.

\subsection{Our contributions}

This paper includes the following main contributions:
\begin{itemize}
\item Provides an asymptotic relative efficiency comparison for GCM and LOCO in 3 different model settings: linear models, additive models and single index models. In particular we prove that GCM is asymptotically more efficient compared to LOCO for the linear model case, provide conditions under which GCM is more asymptotically efficient than LOCO for non-linear models, and single-index models. We further provide a quantity related to an adjusted correlation under which GCM  has superior relative efficiency to LOCO.
\item We also provide a comparison of asymptotic variance between LOCO and more computationally efficient dropout and lazy fine-tuning based estimator~\cite{gao_lazy_2022}.
\item Finally, we conduct a comparison of these different approaches along with other state-of-the-art approaches over a series of simulated and real data with different algorithms including neural networks and tree-based algorithms. The simulation results support the theory and show that GCM is often more accurate although computationally more intensive than other methods.
\end{itemize}

\section{Feature selection method comparison}\label{VI_test}

\subsection{Asymptotic relative efficiency and relative variability}
We begin by discussing how we provide a theoretical comparison for feature selection methods. Since model-agnostic feature selection methods are often difficult to characterize and there are a number of different comparison metric choices, this is a non-trivial issue. In our simulations and real data section, we provide many different comparison metrics.

For the purpose of this paper, we focus on the natural quantity \emph{relative efficiency} for comparing two methods. The next natural question is which statistic we use for the efficiency comparison. When the means of two feature importance estimators are the same, efficiency simply equates to the variance. However, for feature importance statistics across different methods, the variance alone may provide a misleading notion of statistical efficiency, since different methods can produce importance scores on substantially different scales. A statistic with a larger variance may nevertheless exhibit substantially greater discriminatory power if its expected importance under the alternative is correspondingly larger. 

Classical statistics addresses this issue through scale-invariant measures such as the coefficient of variation, which normalizes variability by the magnitude of the mean~\cite{vangel1996confidence}. In the feature selection setting however, the relevant reference point is the expectation under the null hypothesis $H_0$ rather than the absolute mean. This motivates measuring variability relative to the separation between the null and alternative expectations rather than relative to the mean itself. Specifically, let $\mu_0 = \mbox{E}[T \mid H_0]$ and $\mu_1 = \mbox{E}[T \mid H_1]$. We define the \emph{relative variability} of a feature importance statistic as

\[
R(T)
=
\frac{\sqrt{\mathrm{Var}(T \mid H_1)}}
{|\mu_1-\mu_0|}.
\]

Equivalently, its reciprocal

\[
S(T)
=
\frac{|\mu_1-\mu_0|}
{\sqrt{\mathrm{Var}(T \mid H_1)}}
\]

is a standardized signal-to-noise ratio measuring the separation between the null and alternative distributions.

This criterion is closely related to classical standardized effect sizes~\cite{cohen1988statistical} and Pitman efficacy, which characterizes the local asymptotic power of statistical procedures through the ratio of the derivative of the mean to the asymptotic standard deviation~\cite{lehmann2005testing,van2000asymptotic}. Consequently, comparing feature importance statistics through $R(T)$ provides a scale-invariant notion of statistical efficiency that directly reflects their ability to distinguish null from informative features.

\begin{definition}[Relative variability]
Let $T$ be a feature importance statistic with
\[
\mu_0 = \mbox{E}[T \mid H_0], \qquad
\mu_1 = \mbox{E}[T \mid H_1].
\]
We define the \emph{relative variability} of $T$ as
\[
\operatorname{RV}(T)
=
\frac{\sqrt{\operatorname{Var}(T \mid H_1)}}
{|\mu_1-\mu_0|}.
\]
This quantity measures the stochastic variability of the statistic relative to its discriminative signal, namely the separation between its null and alternative expectations.
\end{definition}

In the context of feature selection considered in this paper, the null hypothesis corresponds to zero importance, so that
\[
\mu_0=\mbox{E}[T\mid H_0]=0.
\]
In this case, the relative variability simplifies to
\[
\operatorname{RV}(T)
=
\frac{\sqrt{\operatorname{Var}(T \mid H_1)}}
{|\mbox{E}[T\mid H_1]|}.
\]
Thus, the relative variability is simply the ratio of the standard deviation to the expected importance under the alternative. This naturally leads to the definition of asymptotic relative efficiency when comparing two estimators $T_1$ and $T_2$.

\begin{definition}[Asymptotic relative efficiency]
Let $T^{(n)}_1, T^{(n)}_2$ be two estimators, then the \emph{asymptotic relative efficiency} (ARE) $e_{T_1, T_2}$ based on the relative variability $R(T)$:
$$
e_{T_1, T_2} : = \lim_{n \rightarrow \infty}\frac{\mbox{Var}(T^{(n)}_2)/\mbox{E}^2[T^{(n)}_2]}{\mbox{Var}(T^{(n)}_1)/\mbox{E}^2[T^{(n)}_1]}.
$$
\end{definition}

Next we define the two estimators we focus on in this paper, GCM and LOCO.
\subsection{Generalized Covariance Measure}
GCM is a regression-based conditional independence test. Let's review the setup in Shah
and Peters~\cite{shah_hardness_2020} and connect it to variable selection. Let $(U,V,W) \in \mathbb{R}^{d_U} \times \mathbb{R}^{d_V} \times \mathbb{R}^{d_W}$ with joint distribution $P_0$, where $d_U = d_V =1$ and $d_W \geq 1$. Define:
\[U = f_{0}(W) + \xi \hspace{0.5cm};\hspace{0.5cm} V=g_{0}(W) + \epsilon\] where $f_{0}(W) = E(U|W)$, $g_{0}(W) = E(V|W)$, $\xi = U - f_0(W)$, and $\epsilon = V-g_0(W)$. The generalized covariance between residuals $E[Cov(U,V|W)]$ is defined as: \[\psi^{(gcm)}_{0,(U,V)}:=E[\xi*\epsilon]\]

Given observations $\{(U_i,V_i,W_i)\}_{i=1}^n$, let $\hat f$ and $\hat g$ be estimates of conditional expectations $f_0$ and $g_0$, respectively. The empirical GCM estimator is a normalized version of:
\[
\hat{\psi}^{(gcm)}_{(U,V)}
=
\frac{1}{n}\sum_{i=1}^n
\{U_i-\hat f(W_i)\}\{V_i-\hat g(W_i)\},
\]
which estimates a population residual covariance that is zero under $U\perp V\mid W$.

\begin{theorem}[Re-formulated from \cite{shah_hardness_2020}]
Under suitable assumptions defined in \cite{shah_hardness_2020}, the GCM estimator $\hat{\psi}^{(gcm)}_{(U,V)}$ has the following result: \[\sqrt {n}\big[\hat{\psi}_{(U,V)}^{(gcm)} - \psi_{(U,V)}^{(gcm)}\big] \overset{d}{\rightarrow} N(0, \sigma^2_{(U,V)}), \]where $\sigma^2_{(U,V)} = Var(\xi\epsilon)$.
\end{theorem}

  To see how we apply to the feature selection setting, let's assume we have samples  $Z_i = (X_i,Y_i)$, $i \in [n]$ drawn from data matrices $\boldsymbol{X}^{(n)} \in \mathbb{R}^{n \times p}$ and $\boldsymbol{Y}^{(n)} \in \mathbb{R}^{n }$. Those observations are drawn from a data-generating distribution $P_0$. $X$ (resp. $X_i$) denotes the multivariate random variable containing all features, $Y$ (resp. $Y_i$) denotes the response. For feature index $j \in  [p]$ where $[p]:= \{1,...,p\}$, let $X_j$ (resp. $X_{ij}$) denote the $j^{th}$ feature in $X$ and $X_{-j} \in \mathbb{R}^{p-1}$ (resp. $X_{i,-j}$) denote the features in $X$ with the $j^{th}$ variable removed. In particular, let $(U,V,W)$ be $(X_j,Y,X_{-j})$ applied to the GCM, $\text{for all } j \in [p]$. For notation simplicity we use $\psi_{j}^{(gcm)}$ to represent $\psi_{(X_j,Y)}^{(gcm)} $, then we can test the hypotheses\[H_0:\psi_{j}^{(gcm)} = 0\text{ versus } H_a: \psi_{j}^{(gcm)} \neq 0\ \] We expect the p-value to be small if $Y$ is not conditional independent of  $X_j$  given $X_{-j}$.  

\paragraph{GCM  limitation} 
One of the well-known limitations of GCM is that a covariance of $0$ does not necessarily imply independence which is especially relevant for non-linear relationship. Consider the simple example
$$
Y = X^2 + \epsilon
$$
where $X$ is a symmetric distribution (e.g.normal) and $\epsilon$ is zero-mean independent noise. It follows that $\mbox{Cov}(X^2, Y) = 0$ and the GCM statistic satisfies $\psi_{j}^{(gcm)} = 0$ when clearly $X$ and $Y$ are dependent. This extends to any scenario where the functional relationship is an even function and the random variable has a symmetric distribution. This is a clear limitation of GCM that will be observed in both the theoretical and empirical results.

\subsection{Leave-one-covariate-out}
Before outlining variable importance for feature selection, we review the model-agnostic population framework of Williamson et al.~\cite{williamson_general_2022}, where each feature importance score is measured by the change in predictive accuracy after omitting or replacing the feature.

 Let's denote $s \subset \{1,...,p\}$ as an index set of the covaraite subgroup of interest. Then, $X_{-s} \in \mathbb{R}^{p-|s|}$ under population $P_{0,-s}$ is defined as  the features in $X$ with index in $s$ features removed.  Let $V(f,P_0)$ be a measure of the predictive power of a prediction function $f$ of the model, where $f \in \mathcal{F}$, $\mathcal{F}$ is a rich function class. A natural candidate prediction function would be a population maximizer $f_0$ over the class $\mathcal{F}$:\[f_0 \in \operatorname{argmax}_{\substack{f \in \mathcal{F}}} V(f,P_0).\] Similarly, define prediction function $f_{0,-s}$ be the maximizer over function class $\mathcal{F}_{-s}$ where $\mathcal{F}_{-s}$ is a collection of function $f \in \mathcal{F}$ whose evaluation ignores entries of input $X$ with index in $s$. In the  population setting, for the subset of features $X_s$, importance is estimated by the difference in the predictive power by excluding $s$ features from $X$. Thus, the feature importance score of the subgroup $X_s$ is defined as:\[\psi_{0,s} := V(f_0,P_0) - V(f_{0,-s},P_{0,-s}).\] Throughout this paper we use a negative mean-squared error predictive power:
$$
V(f_0,P_0) = -\mbox{E}[(Y - f_0(X_1, X_2,...,X_p))^2].
$$

For feature selection, we focus on LOCO \cite{lei_distribution-free_2018}, which evaluates one feature at a time by setting $s=\{j\}$ and comparing the full model using $X$ with the reduced model using $X_{-j}$. A feature is considered important if removing it significantly worsens predictive performance. Define
the respective error term as\[\epsilon: = Y- f_0(X)\hspace{0.5cm};\hspace{0.5cm} \epsilon_{-j} := Y-f_{0,-j}(X_{-j})\] where $f_{0}(X) = \mbox{E}[Y|X]$ and $f_{0,-j}(X_{-j}) = \mbox{E}[Y|X_{-j}]$. We can adapt Theorem from  \cite{williamson_general_2022} to LOCO estimator:
 
 \begin{theorem}[Re-formulated from \cite{williamson_general_2022}]
  Under suitable assumptions defined in ~\cite{williamson_general_2022}, the LOCO estimator $\hat{\psi}_{j}^{(loco)} := V(f_n,P_n) - V(f_{n,-j}, P_{n,-j})$ has the following result:\[\sqrt{n}\big[\hat{\psi}^{(loco)}_{j} - \psi_{j}^{(loco)}\big] \overset{d}{\rightarrow} N(0, \sigma^2_j) \] where $\widehat{\psi}^{(loco)}_{j}= \frac{1}{n} \sum_{i=1}^n\{[Y_i-f_n(X_{ij})]^2 - [Y_i - f_{n,-j}(X_{i,-j})]^2\}$, and $\sigma^2_j  =\mbox{Var}(\epsilon^2 - \epsilon_{-j}^2).$
\end{theorem}
Thus, $\text{for all } j \in [p]$,
we can test the hypotheses\[H_0: \psi_{0,j}^{(loco)} = 0\text{ versus } H_a : \psi_{0,j}^{(loco)} \neq 0 .\] We expect the corresponding p-value to be small if $ X_j $ provides additional information for predicting $Y$ not already captured by $X_{-j}$. 

The focus of this paper on the ARE comparison 
$$
e_{\hat{\psi}_{j}^{(gcm)}, \hat{\psi}_{j}^{(loco)}} : = \lim_{n \rightarrow \infty}\frac{\mbox{Var}(\widehat{\psi}^{(loco)}_{j})/\mbox{E}^2[\widehat{\psi}_j^{(loco)}]}{\mbox{Var}(\widehat{\psi}^{(gcm)}_{j})/\mbox{E}^2[\widehat{\psi}_j^{(gcm)}]}.
$$

for these two model-agnostic methods GCM and LOCO. The ARE comparison would apply to any model-agnostic feature importance estimator where the asymptotic mean and variance can be computed. In the simulations and real data comparison, we consider a broader set of methods.

\section{Main results: ARE comparison}\label{sec:models}
In this section, we compare and evaluate the asymptotic relative efficiency of GCM and LOCO for linear, non-linear additive and single index model. We focus on these models because these models allow us to provide theoretical guarantees and they provide a reasonable class of data-generating models. We consider mis-specified models in the simulations section. 

Throughout we define 
$$\tilde{X}_j: = X_j - E[X_j|X_{-j}]$$ 
which denotes the residual based on predicting $X_j$ from $X_{-j}$.  Note that if $X_j$ is independent of $X_{-j}$, then $\tilde{X}_j = X_j - E[X_j]$.

\subsection{Linear Model}
Let's consider $(X,Y) \in \mathbb{R}^p \times \mathbb{R}$ with joint distribution $P_0$ satisfying the linear model \[Y = X_{-j}^T\beta_{-j}+\mathnormal{\beta}_j X_j+\epsilon\]
where $\beta_j \in \mathbb{R}$ and $\beta_{-j} \in \mathbb{R}^{p-1}$ are regression coefficients, and $\epsilon$ is a random noise term with zero mean and finite variance $\sigma^2$.

To select the most important features in the linear model, we are perform individual hypothesis tests: $\text{for all } j \in [p]$,  $H_0: \beta_j =0$ which is equivalent to the test: $X_j$ is dependent of the target $Y$ given the rest features $X_{-j}$, and LOCO test: if $X_j$ provide additional model prediction power when the rest features are already included.

\begin{theorem}\label{thm_linear}
    For a linear model, where $Y = X_{-j}^T\beta_{-j}+\mathnormal{\beta}_j X_j+\epsilon$ and $\epsilon$ is random noise with finite variance $\sigma^2$. $\text{For all } j \in [p]$, we have 
    \begin{equation*}
        \begin{split}
            \frac{sd(\hat{\psi}_{j}^{(gcm)})}{E(\hat{\psi}_{j}^{(gcm)})}&=\frac{\frac{1}{\sqrt{n}}\beta_j\sqrt{Var(\tilde{X}_j ^2) +\frac{1}{\beta_j^2}\sigma^2 E(\tilde{X}_j^2)}}{\beta_j E(\tilde{X}_j^2)},\\
            \frac{sd(\hat{\psi}_{j}^{(loco)})}{E(\hat{\psi}_{j}^{(loco)})}&=
\frac{\frac{1}{\sqrt{n}}\beta_j^2\sqrt{\mbox{Var}(\tilde{X}_j^2)+\frac{4}{\beta_j^2}\sigma^2 E(\tilde{X}_j^2)}}{\beta_j^2 E(\tilde{X}_j^2)}, 
        \end{split}
    \end{equation*}
 and thus $e_{\hat{\psi}_{j}^{(gcm)},\hat{\psi}^{(loco)}_{j}} >1$.
\end{theorem}

The expressions for the ratio of standard deviation and expectation for each estimator is instructive in that they are equivalent once dividing numerator and denominator by $\beta_j$ aside from an additional factor of $4$ in the standard deviation for the LOCO estimator. This displays a clear increase in efficiency for the GCM estimator. As far as we are aware this is the first direct comparison between the two, even for a linear model. And note that there are no assumptions of dependence/correlation so the result holds regardless of the dependence/correlation structure. 

The key quantities that determine how much the efficiency is larger for GCM compared to LOCO is depends on $\frac{\sigma^2}{\beta_j^2}$ as well as $E(\tilde{X}_j^2)$ and $Var(\tilde{X}_j^2)$ where $\tilde{X_j} = X_j - E[X_j|X_{-j}]$. In particular if $\frac{\sigma^2}{\beta_j^2} E(\tilde{X}_j^2)$ is large relative to $Var(\tilde{X}_j^2)$ then the gap in efficiency between GCM and LOCO increases. 

Below is a simple two-variable example illustrating the efficiency ratio.

\begin{example}\label{ex1}
    Suppose $Y = \beta_0+\beta_1 X_1+\beta_2X_2 + \epsilon$ where $X_i \sim N(0,\sigma^2), i=1,2$, $Cov(X_1,X_2) = \rho$, and $\epsilon \sim N(0,\sigma_\epsilon^2)$ is  independent of features. Also, known that $X_1|X_2 \sim N(\rho X_2, (1-\rho^2)\sigma^2)$, then the asymptotic relative efficiency of LOCO and GCM estimators for the testing of $X_1$ is\[e_{\hat{\psi}_{1}^{(gcm)},\hat{\psi}^{(loco)}_{1}} =\frac{4\frac{\sigma_\epsilon^2}{\beta_1^2}+2(1-\rho^2)\sigma^2}{\frac{\sigma_\epsilon^2}{\beta_1^2}+2(1-\rho^2)\sigma^2} >1.\]
    Specifically, if $\frac{\sigma^2_\epsilon}{\beta_1^2} \rightarrow 0$, then $e_{\hat{\psi}_{1}^{(gcm)},\hat{\psi}_{1}^{(loco)}} \rightarrow 1$.
\end{example}

\subsection{Non-linear Additive Model}
\label{subsec: non_linear}

Although the linear model is instructive, the primary focus of this paper is on non-linear additive regression settings.
Let \[Y = \sum_{j=1}^p f_{0,j}(X_j) + \epsilon\] where  $f_{0,j}$'s are non-linear and measurable, and there are no interaction terms among features, $\epsilon$ are i.i.d  mean zero with finite variance $\sigma^2$, and $\epsilon \perp  X$.
For each $j \in [p]$, both GCM and LOCO are performing the following hypothesis test: \[H_0: \psi^{(\cdot)}_{j} = 0 \hspace{0.5cm};\hspace{0.5cm} H_a : \psi^{(\cdot)}_{j} \neq 0.\] 
For regressing  $X_j$ and $Y$ on $X_{-j}$ respectively for GCM, let $g_0,h_0 \in \mathcal{F}$ for function class $\mathcal{F}$, \[X_j=h_{0,j}(X_{-j}) + \xi_{-j} \hspace{0.5cm};\hspace{0.5cm}  Y= g_{0,j}(X_{-j}) + \epsilon_{-j}\] where  $ \epsilon_{-j},  \xi_{-j}$ are i.i.d  mean zero with finite variance $\sigma^2_\epsilon, \sigma_{\xi}^2$ respectively, $\epsilon_{-j},\xi_{-j} \perp  X$. Let
$h_{0,j}(X_{-j}) = E(X_j|X_{-j}), g_{0,j}(X_{-j}) = E(Y|X_{-j})$, recall $\tilde{X_j} : = X_j - E(X_j|X_{-j})$, and $ \tilde{f}_{0,j}(X_j):=f_{0,j}(X_j) - E(f_{0,j}(X_j)|X_{-j})$. We imposing the following assumption on $\mbox{Corr}(\tilde{X}_j, \tilde{f}_{0,j}(X_j))$:

\begin{assumptionp}{(A)}
    $\mbox{Corr}^2(\tilde{X}_j, \tilde{f}_{0,j}(X_j)) >\frac{E(\tilde{X}_j^2 \tilde{f}_{0,j}^2(X_j) )\frac{E(\tilde{f}_{0,j}^2(X_j))}{E(\tilde{X}_j^2)}  + \sigma^2 E(\tilde{f}_{0,j}^2(X_j))}{E(\tilde{f}_{0,j}^4(X_j))+4\sigma^2E(\tilde{f}_{0,j}^2(X_j))}$
\end{assumptionp}

Assumption (A) is an adjusted correlation condition that ensures the efficiency of GCM under ARE comparison. Note that Assumption (A) fails for the example where if $X_j$ is drawn from a symmetric distribution and $f_{0,j}(.)$ is an even function since Corr $(\tilde{X}_j, \tilde{f}_{0,j}(X_j)) = 0$. The assumption therefore ensures that the correlation between $\tilde{X}_j$ and $\tilde{f}_{0,j}(X_j)$ is sufficiently far away from $0$. Also note that in the case of the linear model where $f_{0,j}(X_j) = \beta_j X_j$ Assumption (A) is always satisfied which yields the linear model result from earlier.

\begin{theorem}\label{thm_nonlinear}
    For an additive regression model estimated with mean squared error loss,  $Y = f_0(X) + \epsilon = f_{0,-j}(X_{-j})+f_{0,j}( X_j)+\epsilon$, where $\epsilon$ are i.i.d ransom noise with zero mean and finite variance $\sigma^2$. $\text{For all } j \in [p]$, we have
    \begin{equation*}
        \begin{split}
            \frac{sd(\hat{\psi}_{j}^{(gcm)})}{E(\hat{\psi}_{j}^{(gcm)})}&=\frac{\frac{1}{\sqrt{n}}\sqrt{Var(\tilde{X}_j\tilde{f}_{0,j}(X_j)) + \sigma^2 E(\tilde{X}_j^2)}}{E(\tilde{X}_j\tilde{f}_{0,j}(X_j)) },\\
            \frac{sd(\hat{\psi}_{j}^{(loco)})}{E(\hat{\psi}_{j}^{(loco)})}&=
\frac{\frac{1}{\sqrt{n}}\sqrt{Var(\tilde{f}_{0,j}^2(X_j)) + 4\sigma^2 E(\tilde{f}_{0,j}^2(X_j))}}{E(\tilde{f}_{0,j}^2(X_j)) }.
        \end{split}
    \end{equation*}
    Therefore under Assumption (A),  $e_{\hat{\psi}_{j}^{(gcm)},\hat{\psi}^{(loco)}_{j}} >1.$

\end{theorem}

Next, we present a nonlinear additive regression functions as concrete example to illustrate that the GCM estimator is (approximately) asymptotically more efficient than the LOCO estimator when Assumption (A) in Theorem \ref{thm_nonlinear} is satisfied, and vice versa. The corresponding simulation study will also be shown in Section \ref{sec:simulation}, and another further additive model comparison example can be found in Supplementary Material S1.

\begin{example}\label{ex2}
       Let $Y = \sum_{j=1}^q f_j(X_j) + \epsilon$, where $\epsilon \sim N(0,\sigma^2)$, and $f_j(X_j) = X_j^{p_j}$ where $p_j$ is odd, and suppose $X_j \sim N(0,1)$, and $X_j$'s are independent. Then $\tilde{X_j}  = X_j - E(X_j|X_{-j}) = X_j$. For notation simplicity, we use $\tilde{f}_j$ to represent $\tilde{f}_j(X_j)$, then $ \tilde{f}_{j}=X_j^p - E(X_j^p |X_{-j})= X_j^{p_j}$. 
       
       By the property of  the moment for standard normal that $E(X^{2t}) = \frac{(2t)!}{2^tt!}$ and the sterling approximation $n!\approx \sqrt{2\pi n}(\frac{n}{e})^n$, we have condition (A) in Theorem \ref{thm_nonlinear} as:
    \begin{equation}\label{first}
    \begin{split}
       E(\tilde{f_{j}}^4)\mbox{Corr}^2(\tilde{X}_j, \tilde{f}_j)&+4\sigma^2 \mbox{Corr}^2(\tilde{X}_j, \tilde{f}_j)E(\tilde{f_{j}}^2)\\
        &=E(X_j^{4p_j})\frac{E^2(X_j^{p_j+1})}{E(X_j^{2p_j})} + 4\sigma^2E^2(X_j^{p_j+1})\\
        &\approx \frac{2^{5p_j+1}p_j^{p_j}(p_j+1)^{p_j+1}}{e^{2p_j+1}}+\frac{2^3(p_j+1)^{p_j+1}}{e^{p_j+1}}\sigma^2,
    \end{split}
\end{equation} 
\begin{equation}\label{second}
    \begin{split}
       E(\tilde{X_j}^2 \tilde{f_{j}}^2 )\frac{E(\tilde{f_{j}}^2)}{E(\tilde{X_j}^2)}  + \sigma^2 E(\tilde{f_j}^2) 
        &= E(X_j^{2+2p_j})E(X_j^{2p_j})+\sigma^2E(X_j^{2p_j})\\
        &\approx \frac{2^{2p_j+2}(p_j+1)^{p_j+1}p_j^{p_j}}{e^{2p_j+1}}+\frac{2^{p_j+\frac{1}{2}}p_j^{p_j}}{e^{p_j}}\sigma^2.
    \end{split}
\end{equation}
Thus, (\ref{first}) $>$ (\ref{second}), and $e_{\hat{\psi}_{j}^{(gcm)},\hat{\psi}^{(loco)}_{j}} = \bigg(\frac{sd({\hat{\psi}^{(loco)}_{j}}) / E(\hat{\psi}_{j}^{(loco)})}{sd({\hat{\psi}^{(gcm)}_{j}}) / E(\hat{\psi}_{j}^{(gcm)})}\bigg)^2>1$ up to the Sterling approximation.

\end{example}

Note that in practice, higher-order interaction terms are natural. While our theory does not extend to this setting, we provide a simulation comparison which includes interaction terms in Section~\ref{sec:simulation}.

\subsection{Single Index Model}
\label{subsec: single_index}

Next, we will generalize to a semi-parametric extension of the linear model: single index models which has connections to a shallow neural networks. For single index models with a scalar response variable $Y \in \mathbb{R}$ and a predictor vector $X \in \mathbb{R}^p$, the model is:
\begin{equation*}
    Y = \eta(X^T\beta) + \epsilon
\end{equation*}
where $\beta = (\beta_1, ..., \beta_p)^T$ is unknown parameter vector, $\eta$ is an unknown link function, and $\epsilon$ are i.i.d errors with zero mean and finite variance $\sigma^2$, and $\epsilon \perp X$. Then, for each $j \in [p]$, GCM and LOCO perform the same individual hypothesis testing as discussed previously.
One way to achieve the relative variability of these two estimators in single index model is through Taylor expansion of $\eta(X^T\beta)$ around \(E(X^T \beta | X_{-j}) \). For notation simplicity, we use $\eta' =\eta'(E(X^T \beta | X_{-j}))$.
\begin{equation*}
    \begin{split}
      E(Y|X) = \eta(X^T \beta) &= \eta(E(X^T \beta | X_{-j})) + \eta' \cdot (X^T \beta - E(X^T \beta | X_{-j}))+\Delta
    \end{split}
\end{equation*}
where $\Delta = \sum_{s=2}^{\infty}\frac{1}{s!}\eta^{(s)}(E(X^T\beta|X_{-j}))(X^T \beta - E(X^T \beta | X_{-j}))^s$.
Let's introduce the notation for order of approximation used in the following result: $f(n) = o(1) \text{ means if for all }  \epsilon>0, \text{there exists } M>0, \text{such that } |f(n)| \leq \epsilon \text{ for all } n>M.$
\begin{assumptionp}{(B1)}
    $\eta$ is fixed, monotone differentiable function with bounded derivatives $\eta^{(s)}(X^T\beta)$ $ \leq O(a^{-X^T\beta})$, $ a>1$, for all $s \geq 2$.
\end{assumptionp}
{

\begin{assumptionp}{(B2)}
$X$ has a sub-Gaussian distribution; that is, there exists $\sigma^2>0$ such that
for all vectors $v \in \mathbb{R}^p$ with $\|v\|_2=1$ and all $t \in \mathbb{R}$,
with $T=v^T X$, we have
$P(|T-E(T)|>t) \leq 2\exp\left(-\frac{t^2}{2\sigma^2}\right).$

\end{assumptionp}

\begin{assumptionp}{(B3)}
    $X^T\beta$ has a density that is bounded from above by $\bar{q}$.
\end{assumptionp}

\begin{assumptionp}{(B4)}
For any $\delta>0$,  there exists a constant C, such that $E(X^T\beta|X_{-j}) \geq C \max\{\sqrt\frac{1}{\delta},\sqrt{K}\}$, for some constant  $K\geq 3$ and tail probability $\delta$.
\end{assumptionp}
\begin{theorem}\label{thm_SIM}
   Under the assumptions (B1) to (B4), consider a single index model estimated with mean squared error loss,  $Y = \eta(X^T\beta)+\epsilon$, where $\epsilon$ is random noise with finite variance $\sigma^2$. $\text{For all } j \in [p],$ we have 
   \begin{equation*}
       \begin{split}
           \frac{sd(\hat{\psi}_{j}^{(gcm)})}{E(\hat{\psi}_{j}^{(gcm)})}&=\frac{\frac{1}{\sqrt{n}}\sqrt{\eta'^2 \beta_j^2Var(\tilde{X}_j^2)+ \sigma ^2 E(\tilde{X}_j^2 ) + o(1)}}{\beta_j \eta'E(\tilde{X}_j^2)+o(1)};\\
           \frac{sd(\hat{\psi}_{j}^{(loco)})}{E(\hat{\psi}_{j}^{(loco)})}&=\frac{\frac{1}{\sqrt{n}}\sqrt{\eta'^4\beta_j^4 Var(\tilde{X}_j^2)+4\sigma^2\eta'^2\beta_j^2E(\tilde{X}_j^2)+o(1)}}{\beta_j^2 \eta'^2 E(\tilde{X}_j^2)+o(1)}.
       \end{split}
   \end{equation*}
   For any $\epsilon' >0, \delta>0$, with probability at least $1-\delta$,  we have $e_{\hat{\psi}_{j}^{(gcm)},\hat{\psi}^{(loco)}_{j}} >1-\epsilon'.$
\end{theorem}
\begin{remark}
\end{remark}
\begin{itemize}
    \item \textit{The asymptotic efficiency proof technique here is similar to Theorem \ref{thm_linear} except that we additionally employ a linear approximation via Taylor series expansion of $\eta(X^T\beta)$ around $E(X^T \beta | X_{-j})$, so assumption (B1) and (B4) ensure that first order approximation of $\eta$ is sufficient and higher order derivatives terms are negligible, such that $\Delta, E(\Delta|X_{-j}) =o(1)$.}
    \item \textit{(B2) and (B3) are relaxed assumptions for the guarantee of the convergence of the least-square estimator of $(\eta, \beta)$ in monotone single index models introduced in Theorem 4.1 in ~\cite{balabdaoui_least_2019}. Those assumptions will also be used to bound the polynomial terms in Taylor series expansion.}
\end{itemize}

The single index model is useful given its connection to single-layer neural networks if we have an unknown and fixed link function, or a generalized linear model if the link function is known. Next, we illustrate a binary classification example to compare the relative efficiency of GCM and LOCO hypothesis testing where we continue using the same linear predictor set up as in Example \ref{ex1}  but with the sigmoid link function.
\begin{example}
    Suppose $Y = \eta(\beta_0+\beta_1 X_1+\beta_2X_2) + \epsilon$ where $\eta$ is sigmoid function (ie: $\eta(x) = \frac{1}{1+e^{-z}}$), $X_i \sim N(0,\sigma^2), i=1,2$, $Cov(X_1,X_2) = \rho$, and $\epsilon \sim N(0,\sigma_\epsilon^2)$ is independent of matrix $X$.
    
    The sigmoid function $\eta$ is an increasing function with recursive derivatives~\cite{minai_derivatives_1993} $\eta^{(s)}(x) = \sum_{k=0}^{s-1} (-1)^k \left\langle \genfrac{}{}{0pt}{}{s-1}{k} \right\rangle \eta(x)^{k+1} (1 - \eta(x))^{s-k}$,
where $\left\langle \genfrac{}{}{0pt}{}{s-1}{k} \right\rangle$ are Eulerian numbers. Thus, $ \eta^{(s)}(x) \leq e^{-|x|} C_s = O(e^{-|x|}) = o(1)$ for $s\geq 2$ when $|x| \rightarrow \infty$, so the higher order derivatives in the remainder term of linear Taylor series expansion is negligible. The polynomial term in the remainder decays fast because of normal random variable $X_j$. For notation simplicity, let $\eta'=\eta'(E(\beta_0+\beta_1 X_1+\beta_2X_2|X_2)) =\eta'(E(\beta_0 + \beta_1 \rho X_2 + \beta_2 X_2))= \eta\left(\beta_0 + (\beta_1 \rho + \beta_2) X_2\right) \cdot \left(1 - \eta\left(\beta_0 + (\beta_1 \rho + \beta_2) X_2\right)\right)$.
    Thus, the relative efficiency of GCM and LOCO test statistics for $X_1$ is\[e_{\hat{\psi}_{0,1}^{(gcm)},\hat{\psi}_{0,1}^{(loco)}} = \bigg(\frac{sd({\hat{\psi}^{(loco)}_{0,1}})}{\beta_1 \eta'sd({\hat{\psi}^{(gcm)}_{0,1}})}\bigg)^2=\frac{\frac{4}{\beta_1^2\eta'^2}\sigma_\epsilon^2(1-\rho^2)\sigma^2+2(1-\rho^2)^2\sigma^4}{\frac{1}{\beta_1^2\eta'^2}\sigma_\epsilon^2(1-\rho^2)\sigma^2+2(1-\rho^2)^2\sigma^4}\bigg(1+o(1)\bigg).\]
Thus, for any $\epsilon' >0$, there exists a  sufficiently large $\eta'$, such that  we have $e_{\hat{\psi}_{0,1}^{(gcm)},\hat{\psi}^{(loco)}_{0,1}} >1-\epsilon'$.
\end{example}

\section{Extension to comparison with Dropout and Lazy-VI}\label{sec:lazy}
The previous section compares the asymptotic relative efficiency for GCM and LOCO. In this section, we extend the comparison to estimators from dropout and Lazy Variable Importance (Lazy-VI)~\cite{gao_lazy_2022} approaches which are used as computationally faster approaches that attempt to estimate similar parameters to LOCO. Let's first review the setup of dropout and Lazy-VI.
\subsection{Dropout}
The dropout approach is a counterpart to LOCO which avoids retraining a reduced model. Instead, the $j^{th}$ feature is replaced by its corresponding marginal mean and then reinserted into the pre-trained full model $f_0$. Let's define the dropout features as $X^{(j)}$ (resp. $X_i^{(j)}$), where $X^{(j)}= (X_1,...,X_{j-1},\mu_j, X_{j+1},...,X_p)$. The plug-in estimator for the dropout is \[\hat{\psi}^{(dr)}_{j} := V(f_n,P_n) - V(f_{n},\tilde{P}_{n,-j})\] where $\tilde{P}_{n,-j}$ is the empirical distributions of $X^{(j)}$. The dropout estimator for feature $j$ is :  \[\widehat{\psi}^{(dr)}_{j}= \frac{1}{n} \sum_{i=1}^n\{[Y_i-f_n(X_{ij})]^2 - [Y_i - f_{n}(X_i^{(j)})]^2\}.\]

\subsection{Lazy-VI}
The prediction functions in GCM, LOCO, and dropout tests can be estimated using user-selected learners. Lazy-VI \cite{gao_lazy_2022} is a computationally efficient approximation to LOCO, primarily developed for neural networks because it relies on gradient-based updates, and an ensemble-based algorithm like random forest does not apply. Instead of retraining a reduced model for each feature, Lazy-VI fits the full neural network once and approximates the reduced model by solving a linearized ridge regression around the full-model parameters. Thus, Lazy-VI reduces the computational cost of LOCO while retaining asymptotic guarantees under neural network setup. We briefly outline its setup below.

 Let a neural network function class $\left\{h_\theta(x): \mathbb{R}^p \mapsto \mathbb{R} \mid \theta \in \mathbb{R}^M\right\}$ be parameterized by a vector $\theta$, such that the full model estimator is
$$
\theta_f=\underset{\theta \in \mathbb{R}^M}{\arg \min } \frac{1}{n} \sum_{i=1}^n\left[Y_i-h_\theta\left(X_i\right)\right]^2.
$$
However, unlike LOCO, which need to retrain the reduced model,
 Lazy-VI approximates the reduced-model parameter by $\theta_f+\Delta\theta_j(\lambda,n)$, where
\begin{align*}
\Delta \theta_j(\lambda, n)=\underset{\omega \in \mathbb{R}^M}{\arg \min } & \left\{\frac { 1 } { n } \sum _ { i = 1 } ^ { n } \left[Y_i-h_{\theta_f}(X_i^{(j)})\right.\right. \left.\left.-\left.\omega^{\top} \nabla_\theta h_\theta(X_i^{(j)})\right|_{\theta=\theta_f}\right]^2+\lambda\|\omega\|_2^2\right\}
\end{align*}
and $\lambda>0$ is the penalty parameter.
Hence, the variable importance score under Lazy-VI is
$$
\psi_{j}^{(lazy)}=V\left(h_{\theta_f}, P_0\right)-V\left(h_{\theta_f+\Delta \theta_j(\lambda, n)}, P_{0,-j}\right).
$$

\begin{theorem}\label{thm_lazy}(Re-formulated from \cite{gao_lazy_2022})
Under the regularity conditions required in Theorem 4.4 of \cite{gao_lazy_2022}, the Lazy-VI estimator under the mean squared error predictive measure
\[
\widehat{\psi}_{j}^{(lazy)}=\frac{1}{n} \sum_{i=1}^n\left\{\left[Y_i-h_{\theta_f}\left(X_i\right)\right]^2 - \left[Y_i-h_{\theta_f+\Delta \theta_j(\lambda, n)}(X_i^{(j)})\right]^2\right\}
\] has the following result:
\[\sqrt{n}[\hat{\psi}^{(lazy)}_{j} - \psi_{j}^{(lazy)}] \overset{d}{\rightarrow} N(0, \sigma^2_j) \text{, where $\sigma^2_j  = Var(\epsilon^2 - \epsilon_{-j}^2).$ }\]
\end{theorem}

LOCO, dropout and Lazy-VI are asymptotically unbiased estimators used to estimate the ability of $X_j$ in predicting $Y$ in terms of the variance only. Then, we can extend the asymptotic relative efficiency comparison to those methods in the following theorem: 

\begin{theorem}\label{thm_dr_loco}
    Under the regularity conditions required in Theorem 4.4 of \cite{gao_lazy_2022}, together with respective assumptions in Theorem \ref{thm_linear} Theorem \ref{thm_nonlinear}, and Theorem \ref{thm_SIM}, then we have the corresponding asymptotic variance relationship:
    \[\lim_{n\to\infty}Var(\hat{\psi}_{j}^{(loco)}) {=} \lim_{n\to\infty}Var(\hat{\psi}_{j}^{(lazy)}){\leq} \lim_{n\to\infty}Var(\hat{\psi}_{j}^{(dr)}).\]    
\end{theorem}

\section{Simulations}\label{sec:simulation}
In this section, we validate our theoretical results by comparing GCM, LOCO, dropout and Lazy-VI. We also provide a comparison to other state-of-the-art feature selection methods such as Lasso~\cite{ma_penalized_2008}, Knockoff~\cite{barber_controlling_2015, candes_panning_2018}, and permutation testing~\cite{breiman_random_2001}. We assess their efficiency and accuracy for the linear model, single index model, and various non-linear models both with and without interaction terms. Specifically, the following simulation settings are used:
\begin{enumerate}
    \item[(a)] Linear model:\\
    $Y \sim 0.01X_1+0.1X_2+X_3+X_4+1.5X_5+2X_6+3X_7+4X_8+\epsilon$, where $X \sim N(0,\Sigma_{20 \times 20})$, $ \text{Corr}(X_3,X_4) = 0.5$, $\epsilon \sim N (0, 0.1^2).$
            \item[(b)] Non-linear additive model  :\\
            $ Y \sim 2X_1^2 + 2\cos(4X_2) + \sin(X_3) + \exp\left(X_4/3\right) + 3X_5 +X_6^3 + 5 X_7 + \max(0, X_8) +\epsilon$, where $X \sim N(0,\Sigma_{20 \times 20}), \epsilon \sim N (0, 0.1^2)$;
        \item[(c)] Non-additive model with interaction :\\
        $Y \sim 2\sin(X_1) + \log(|X_2| + 1) + 3X_1X_2 + \cos(X_3+X_4) + X_5^3 + X_6X_7X_8+ \epsilon$,  where $X \sim N(0,\Sigma_{20 \times 20}), \epsilon \sim N (0, 0.1^2)$;
        \item[(d)] Single Index Model :\\
        $Y \sim f(X\beta)+\epsilon$, where $f$ is sigmoid function, $\beta  = (6,5, 4, 3, 2, 1,0.5,0.1,0,...,0)^T \in \mathbb{R}^{20}, X \sim N(0,\Sigma_{20 \times 20})$, $ \text{Corr}(X_1,X_2) = 0.5, \epsilon \sim N (0, 0.1^2)$.
\end{enumerate}

\subsection{Variable selection comparison across methods}
In this section, the key theoretical results will be validated by the following simulations. For all models (a)-(d), the response variable depends only on the first $8$ of $p=20$ features, and we use $n=1000$ observations in each of the $100$ simulations. We discuss the feature selection methods through canonical implementations designated for each model (see Supplementary Material S4 for more details). However, the variable importance selection methods we investigate here are not restricted to regression problems. We consider both algorithms suggested by the corresponding models, and the same method across different models.
\begin{figure}[t!]
    \centering
        \includegraphics[width=0.5\textwidth]{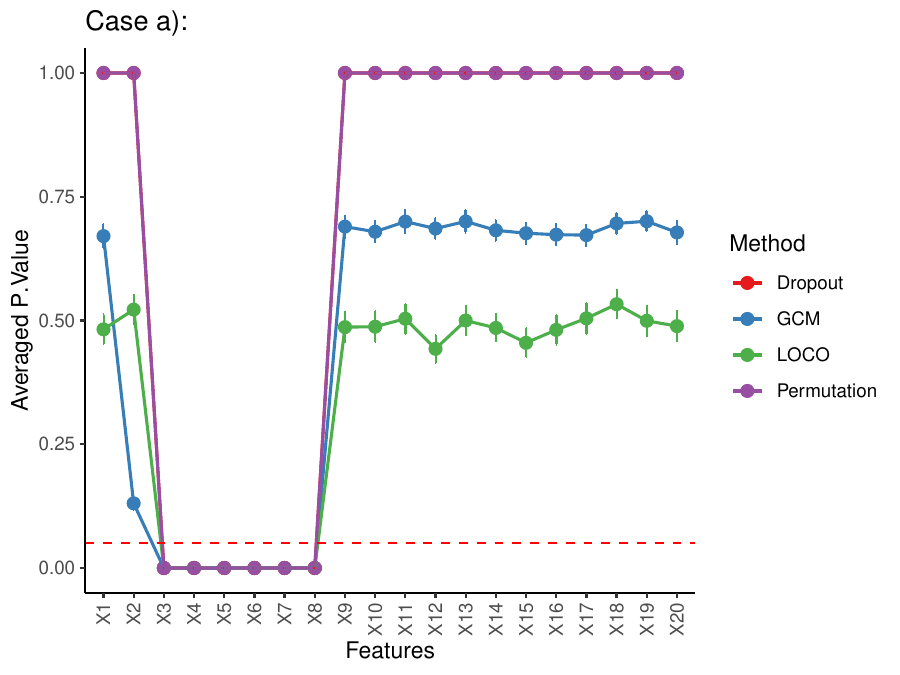}%
    \hfill
        \includegraphics[width=0.5\textwidth]{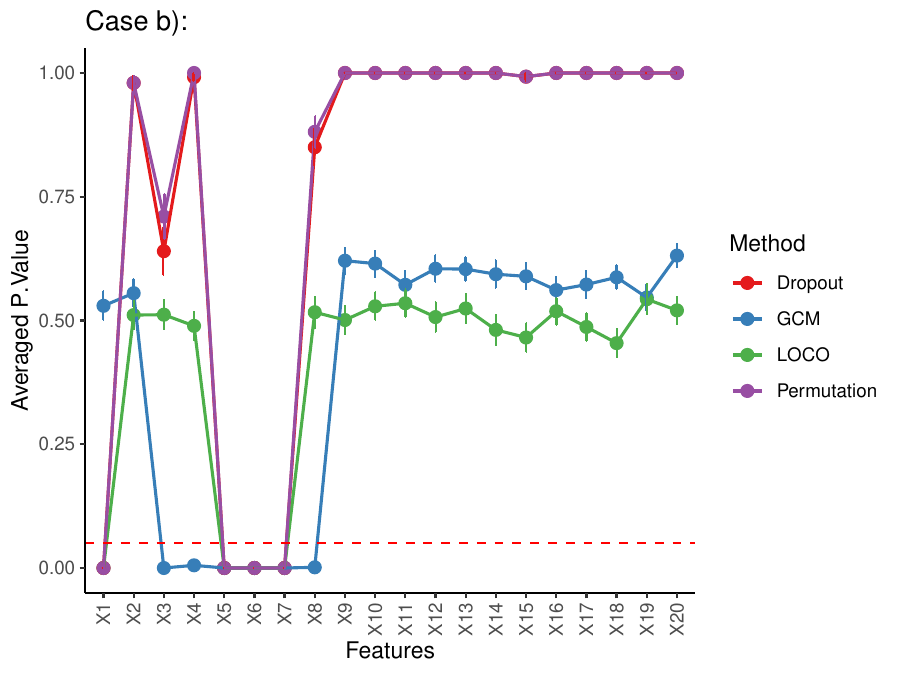}%

    \vspace{0.1cm}
        \includegraphics[width=0.5\textwidth]{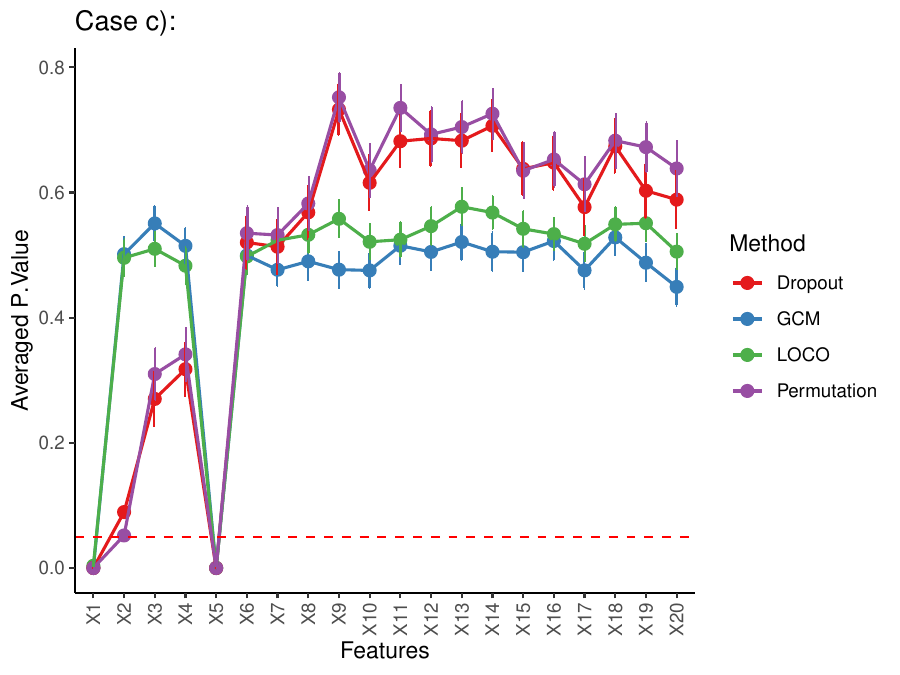}%
    \hfill
        \includegraphics[width=0.5\textwidth]{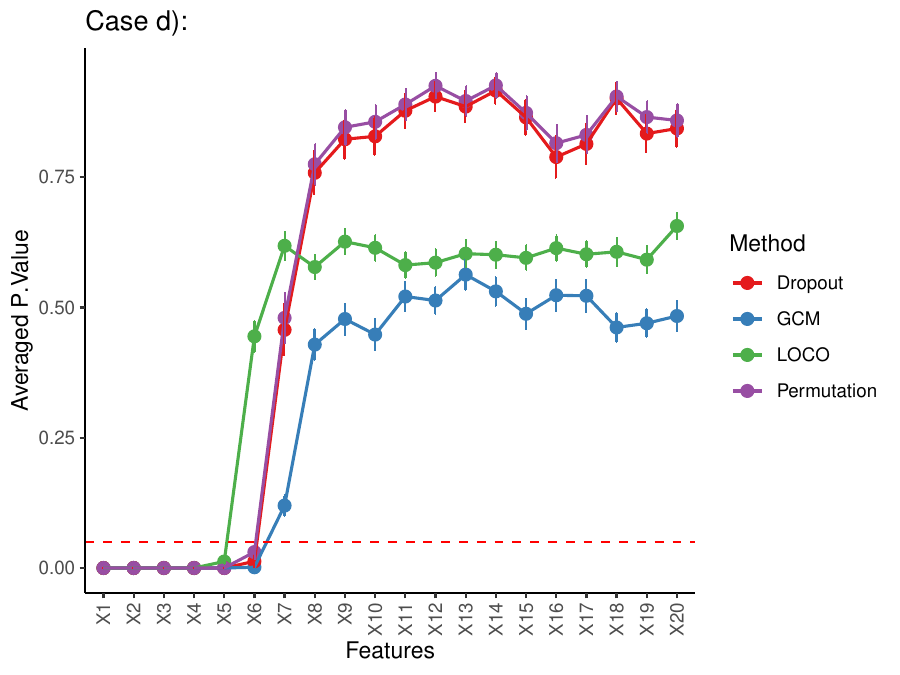}%

       \caption{Averaged p-value and standard deviation of each feature for model (a) - (d)  for GCM, LOCO, dropout and permutation univariate selection on dataset using gradient boosted decision tree. Red dashed line: targeted type-I error rate ($\alpha = 0.05$). }
    \label{gcm_loco_gbm_plots}
\end{figure}
 Specifically, we choose a widely used tree-based algorithm: gradient boosted decision tree with default setting (\texttt{gbm}, \cite{ridgeway_gbm_2024} package in \texttt{R}). The average p-value for each feature performed by GCM, LOCO, dropout and permutation variable selection is presented in Fig.~\ref{gcm_loco_gbm_plots}. What's more, the performance comparison is also extended with  other popular methods which do not provide p-values, e.g. Lasso and Knockoff with target FDR = 0.2 across all models, is reported in Fig.~\ref{performance_metric_plot} and the dashed line is the empirical threshold from random coin-flipping. (\texttt{glmnet}, \cite{friedman_glmnet_2023}; \texttt{knockoff}, \cite{barber_knockoff_2022} packages in \texttt{R}).

\begin{figure}[t!]
\centering
\includegraphics [scale=0.7]{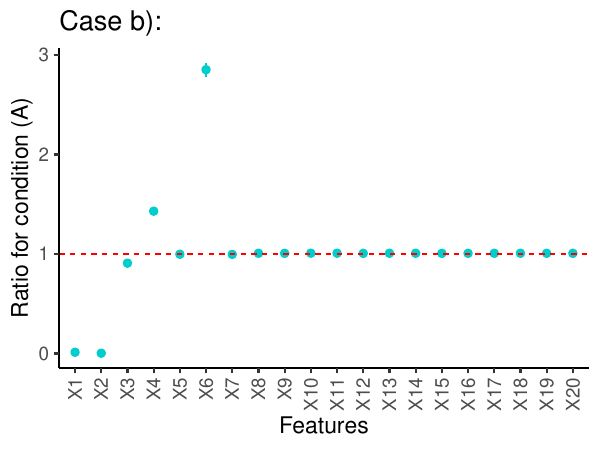}
     \caption{Averaged ratio with standard deviation for verifying theoretical condition (A) (LHS/RHS) in Theorem \ref{thm_nonlinear} for case b).}
    \label{ratio_b_plt}
\end{figure}
\begin{figure}[t!]
  \centering
  \includegraphics[width=0.9\linewidth]{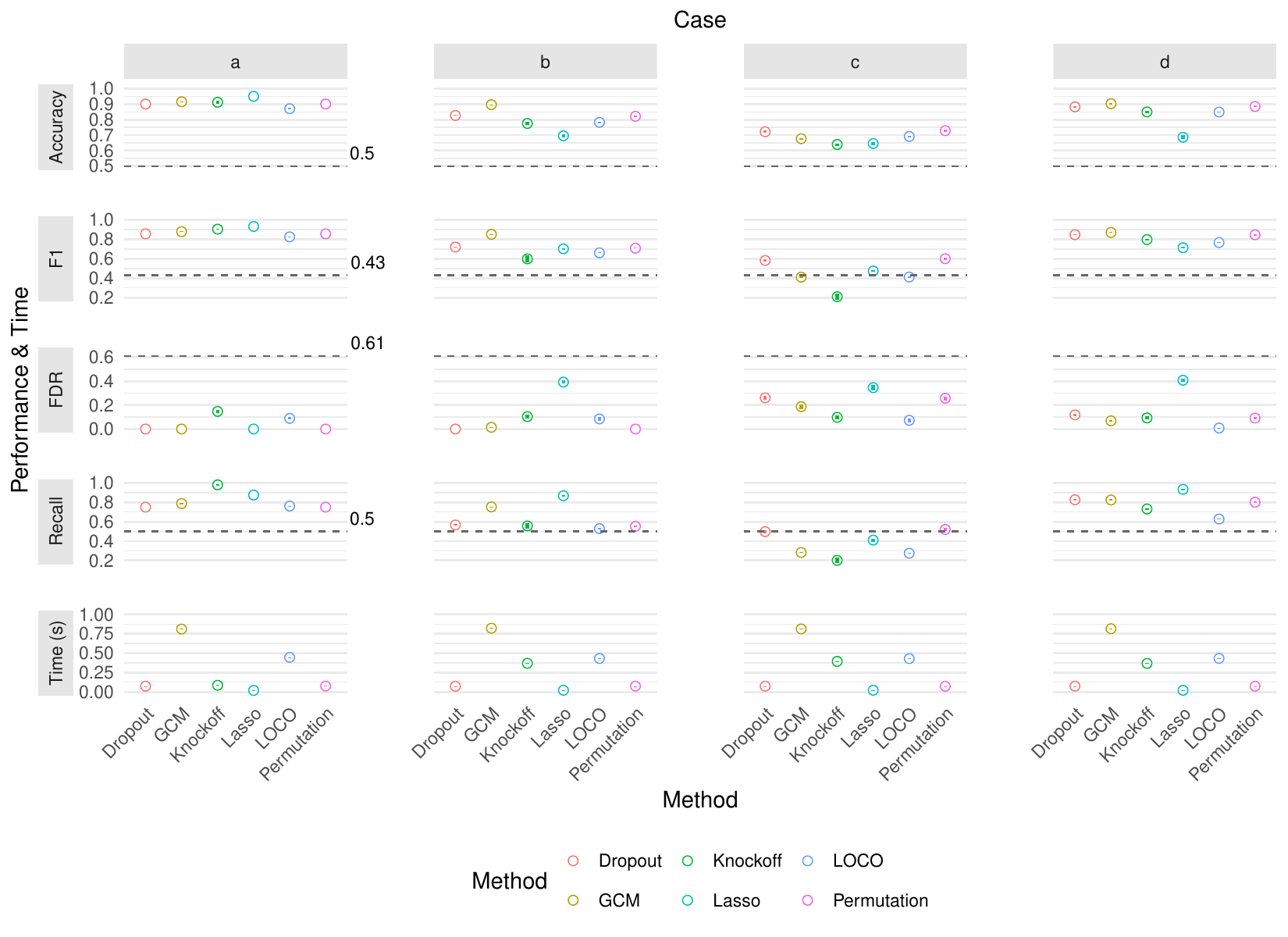}
  \caption{Average performance measure and time (seconds) with standard deviation across 100 simulations for lasso, knockoff (target FDR = 0.2), GCM, LOCO, dropout and permutation univariate selection on dataset using gradient boosted decision tree for (a) - (d). Dashed line: empirical random‐guessing baseline.}
  \label{performance_metric_plot}
\end{figure}
In Fig.~\ref{gcm_loco_gbm_plots}, plot (a) shows that all approaches have similar performance in the linear model case. The performance of lasso is slightly better than others with the highest averaged accuracy and F1 score as shown in Fig.~\ref{performance_metric_plot}. For the additive non-linear model (b) in Fig.~\ref{gcm_loco_gbm_plots}, LOCO requires approximately half the computational time of GCM, but GCM can identify more significant variables than LOCO as long as the ratio checking (see Fig.~\ref{ratio_b_plt}) LHS over RHS in assumption (A) in Theorem \ref{thm_nonlinear} holds. The first two ratios are 0 because the quadratic and cosine functions are even functions and yield the corresponding $\mbox{Corr}(\tilde{X}_j, \tilde{f}_{0,j}(X_j))=0, j=1,2$, which violates assumption (A). What's more, GCM outperforms knockoff and lasso methods in the non-linear model with high accuracy, F1 score and low FDR, since the underlying model is not sparse as shown in Fig.~\ref{performance_metric_plot}. The simulation with the non-additive nonlinear model (c) examines the necessity of assuming the absence of joint effects in the application of all variable selection approaches even under ``black-box'' machine learning algorithm, since all of the approaches barely identify the significant features which yields low recall. For single index model (d), as illustrated in Fig.~\ref{gcm_loco_gbm_plots}, all four methods effectively identify variables with relatively large weights but fail to detect those with smaller contributions. The GCM measure is the most computationally demanding method, as it requires training $2p$ models. In contrast, the LOCO approach trains $p+1$ models, while the dropout, permutation and lasso method requires less computational effort with only a single model needs to be trained. The knockoff filter method maintains single-model computation efficiency but requires extra time for knockoff feature generation. Overall, the GCM approach is more stable and effective compared to all other feature selection methods. Additional simulation results for models trained with kernel SVM, presented in Supplementary Material S4, further validate the strength and robustness of GCM.

\subsection{Model Mis-specification}
To assess the robustness of the proposed comparisons beyond the correctly specified settings used in the theoretical results, we consider two mis-specified models, one with interactions and the other with heavy-random variable with heavy-tailed distributions (which violates sub-Gaussianity) with $n = 1000$ observations in each of the $50$ simulations:
\begin{enumerate}
    \item[(e)] Non-additive model with surrogate feature:\\
    $Y \sim \beta X_1 + 0.05\left(X_3X_4^2 + X_4X_3^2\right) + X_5 + \varepsilon, \quad \varepsilon \sim N(0,1)$, and $X_2 = \rho X_1 + \sqrt{1-\rho^2}Z$, where $Z \sim N(0,1), \rho \in \{0, 0.3, 0.5, 0.7, 0.8, 0.9\}$ and $X \sim N(0,I_{20 \times 20})$.
            \item[(f)]Nonlinear additive model with heavy-tailed covariates  :\\
            $ Y \sim \beta X_1 + 1.5\sin(X_2) + 2*\mathbf{I}\{X_3 > 0\} + \tanh(X_4) + 0.6X_5^3 +\epsilon, \text{where } \epsilon \sim N(0,1), 
 X_{j}\overset{\mathrm{i.i.d.}}{\sim}t_\nu, \nu \in \{3, 5, 10,15, 30, \inf\}, j =1,...,20$.
\end{enumerate}
\begin{figure}[htbp]
    \centering
    \includegraphics[width=0.83\linewidth]{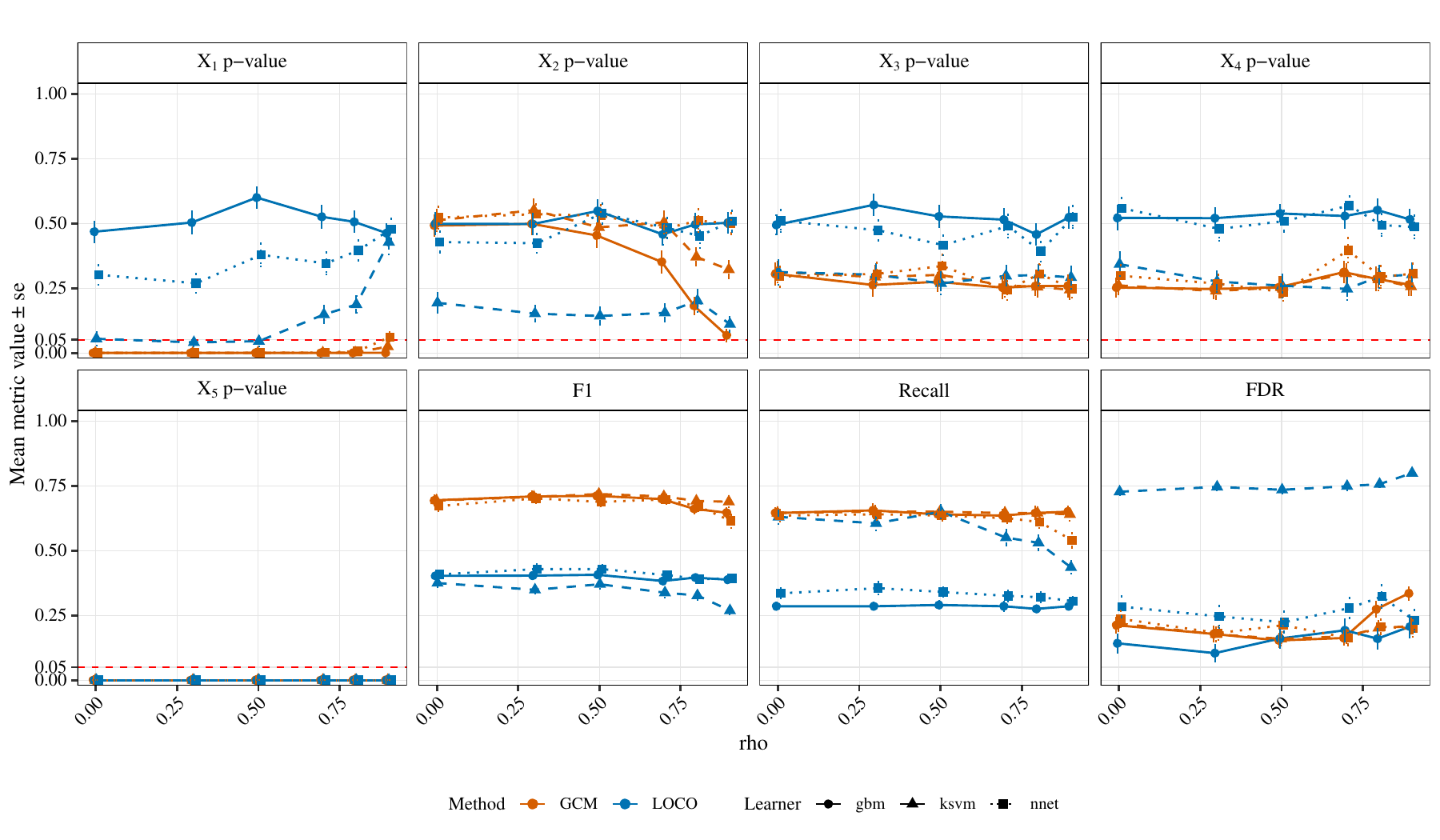}
    \caption{Performance of GCM and LOCO under $t$-distributed covariates with varying degrees of freedom. Panels show mean $X_j$ p-values and performance metrics over 50 simulations, with error bars denoting standard errors. The blue dashed line marks $\alpha = 0.05$.
}
    \label{mis_rho}
\end{figure}
In model (e), $X_2$ is correlated with the significant variable $X_1$, but has no direct effect on the response. The model is intended to evaluate robustness of comparison between LOCO and GCM to model misspecification. For the learners, GBM is implemented with 50 trees and a learning rate of 0.05, neural networks are fitted using the default hyperparameters, and KSVM is fitted with (C=0.1). From Fig.\ref{mis_rho}, we see that across different learners, conclusions from the theoretical results remain consistent. GCM always identifies $X_1$ regardless of whether $X_2$ is weakly or strongly correlated with $X_1$, whereas LOCO is more sensitive to the correlation and produces more false discoveries. Neither method reliably detects the non-additive interaction term $X_3$ and $X_4$. Overall, GCM is more stable than LOCO, achieving higher F1 score and recall.
\begin{figure}[htbp]
    \centering
    \includegraphics[width=0.83\linewidth]{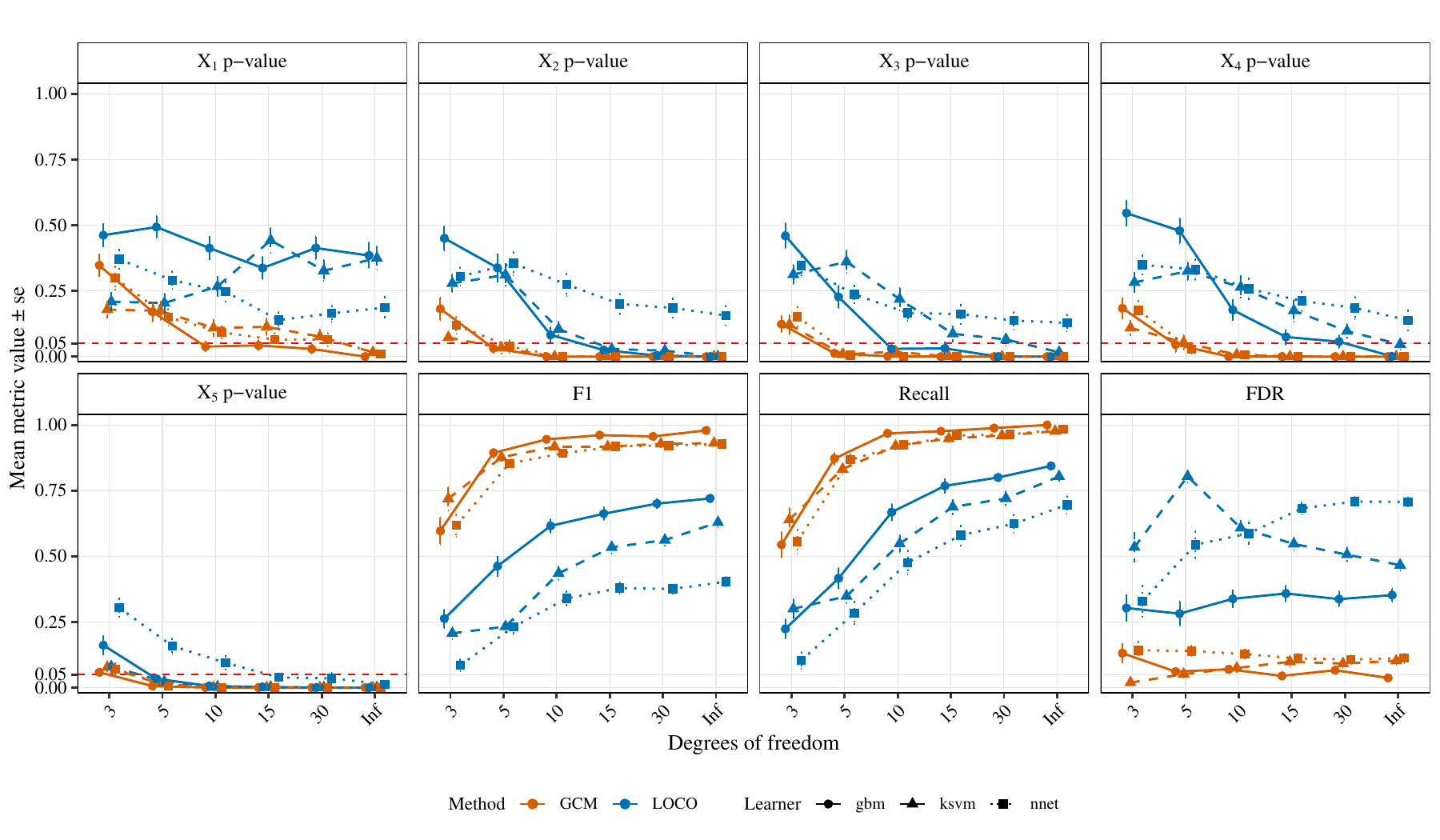}
\caption{
Performance of GCM and LOCO under $t$-distributed covariates with varying degrees of freedom. Panels show mean $X_j$ p-values and performance metrics over 50 simulations, with error bars denoting standard errors. The blue dashed line marks $\alpha = 0.05$.
}
    \label{mis_dft}
\end{figure}
Model (f) considers a nonlinear additive model with heavy-tailed covariates. The covariates are generated independently from t-distributions. To make it comparable across different degrees of freedom, each covariate is standardized to have unit variance. The model evaluates the performance of GCM and LOCO generalizes beyond the model from beyond Gaussian or light-tailed covariate distributions. All three learners are implemented with default settings. Fig.\ref{mis_dft} shows that GCM is more stable than LOCO across different degrees of freedom and machine learners. GCM consistently achieves higher F1 score and recall while maintaining a lower FDR. In contrast, LOCO performs worse under heavier tails, but gradually improves as the degrees of freedom increase, in other words, the covariate approaches Gaussian distribution, allowing it to better identify the significant features.
 
These findings indicate that the superior performance of GCM remains across the misspecified settings considered, suggesting greater empirical robustness beyond standard modeling assumptions.

\subsection{Comparison with Lazy-VI in neural networks}\label{comp:lazy}
In the next simulation, we extend the performance comparison with Lazy-VI in a wide neural network setup.  For this comparison, we implement simulations in Python as the estimation procedure for a lazy training method is not easily transferable to
R because it relies on a sufficiently large neural network function class and other designated assumptions. To generate synthetic data, we use the same setup of the neural network as that of \cite{gao_lazy_2022} proposed for evaluating the performance of Lazy-VI. We train a wide,
fully connected two-layer neural network for all simulations, using Adam with learning rate $10^{-3}$ and the width of the hidden layer in the training network is $50$.

\begin{figure}[t]
    \centering
    \includegraphics[width=0.9\linewidth]{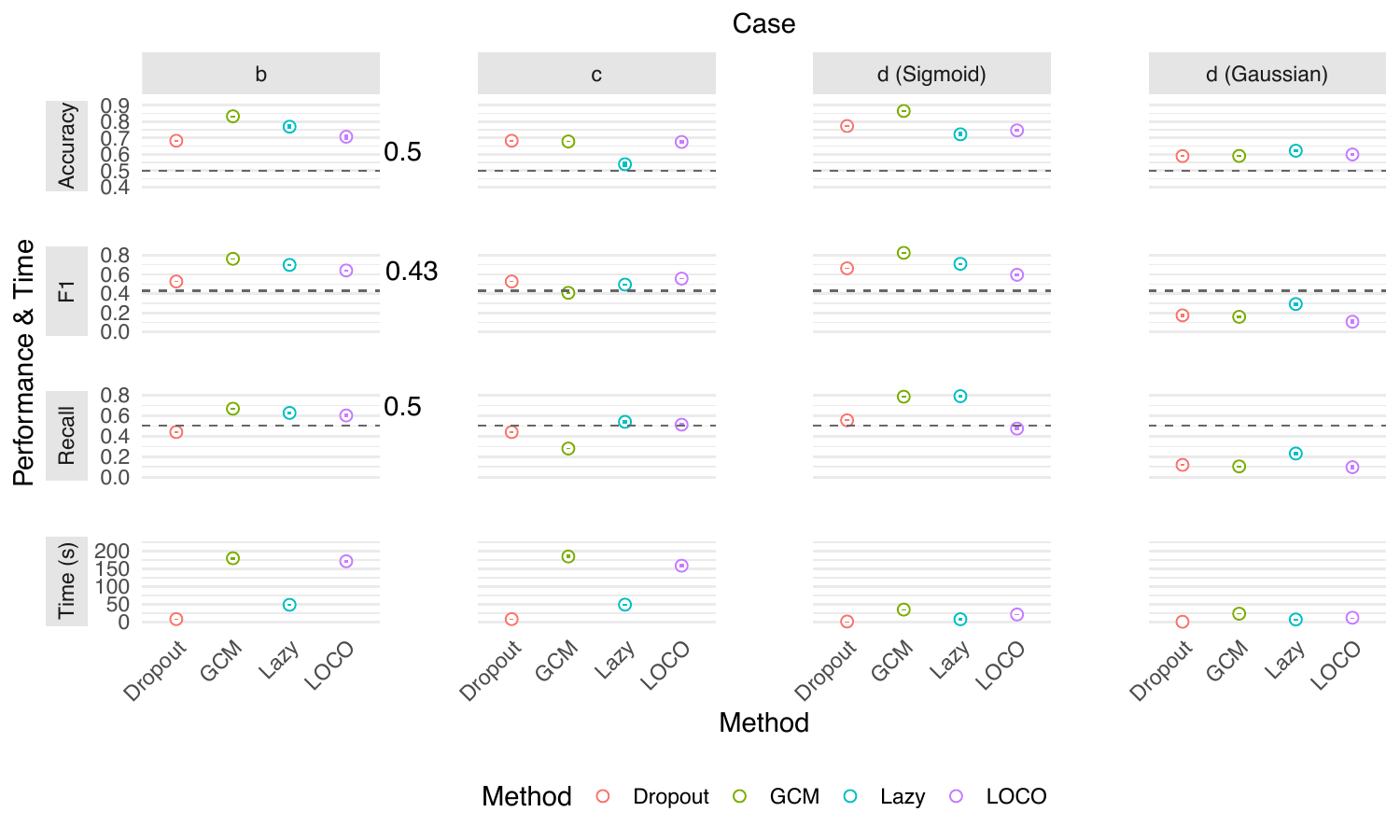}
    \caption{Average performance measure and time (seconds) with standard deviation across 100 simulations for GCM, LOCO, dropout and Lazy-VI univariate selection on dataset using large neural network for (b), (c), and (d) with sigmoid and gaussian link function. Dashed line: empirical random‐guessing baseline.}
    \label{perf_py}
\end{figure}

In Fig.~\ref{perf_py}, we display the performance comparison among LOCO, GCM, Lazy-VI and dropout across 100 simulations. We find that GCM still has better performance than others in terms of high accuracy, F1 score and recall in model b) and d) with sigmoid link function albeit at the expense of computational time. Lazy-VI has proved to be fast and accurate in approximating the LOCO test. LOCO and Lazy-VI have similar performance which is consistent with our theory. For model (c), all methods perform poorly due to interaction terms, resulting in low statistical power. For the single index model (d), when the monotonicity assumption (B1) in Theorem \ref{thm_SIM} is violated,  the gaussian link function (ie: $e^{-x^2}$), none of the feature selection methods work well.

\subsection{High-dimensional Regression}\label{sec:HD}
We also check the behavior of all feature selection methods in high-dimensional setting. 
We use $500$ covariates. Specifically, the first $40$ variables are significant and are generated by repeating the same eight-significant-feature block in model (a) and (b) five times, while the remaining $460$ variables are noise variables. In setting (a), all covariates are independent except that the third and fourth active variables within each repeated block have correlation $0.5$. In setting (b), all covariates are independent. The computational burden is more pronounced in high-dimensional setup since GCM requires retraining more models than other methods.  Fig.~\ref{performance_metric_HD} displays the average performance of each feature selection method over 50 simulations in the high-dimensional setting. All methods are implemented using gradient boosted trees with the default \texttt{gbm} settings in R, except that the tree depth is set to 4. The dashed line represents the empirical threshold obtained from fair coin-flipping sampling. Knockoff filter method works well in all sparse models with high F1 score and low FDR, lasso has the best performance in sparse linear model (a). The performance of GCM is slightly better than other wrapper-based selection with higher accuracy, recall and F1 score in case (a), and GCM performs as good as others in non-linear high-dimensional setting. Model training choices also influence feature selection. Further simulation results with kernel SVM also validate the stability and power of GCM (see Supplementary Material S4).
\begin{figure}[t!]
  \centering
  \includegraphics[width=0.9\textwidth]{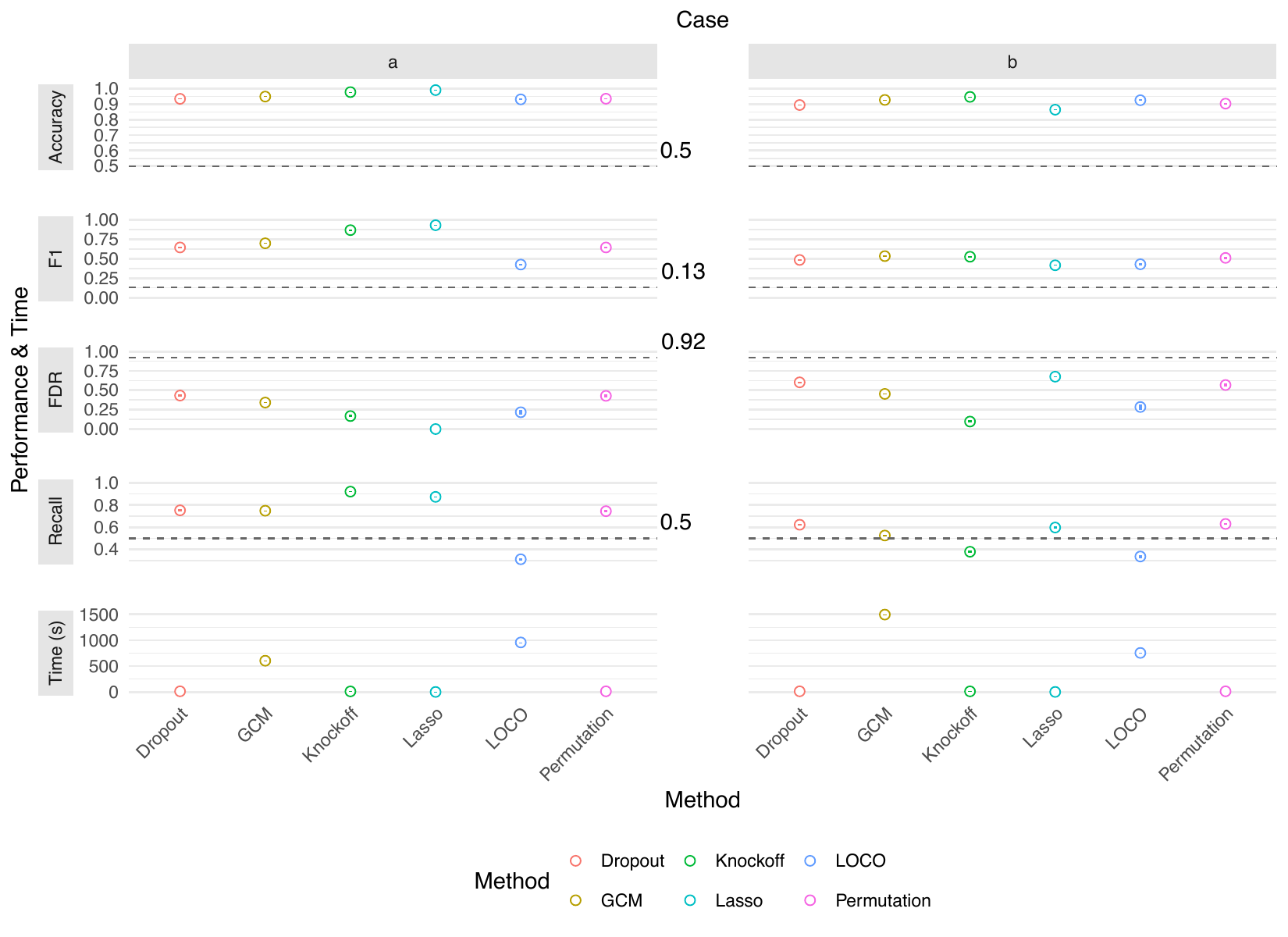}
  \caption{Average performance measure and time (seconds) with standard deviation across 50 simulations for lasso,  knockoff (target FDR = 0.2), GCM, LOCO, dropout and permutation univariate selection on dataset using GBM for (a)-(d). Dashed line: empirical random‐guessing baseline. }
  \label{performance_metric_HD}
\end{figure}

\section{Real data analysis}\label{sec:real_data}
 
In this section, we evaluate GCM and LOCO on two real datasets: Airbnb pricing data~\cite{insideairbnb2026} and social media addiction data~\cite{asif_zaman_2026}. We use 5-fold cross-validation and report the average MSE with its standard error across folds in tables. Stability selection~\cite{meinshausen_stability_2010} where if the feature is selected in $4$ or $5$ folds is used to select features. We use linear regression, gradient boosted decision tree and neural network prediction algorithms. Additional data analysis results are available in Supplementary Materials S4.

\subsection{Airbnb listing price data}
As Airbnb has transformed the short-term rental market, understanding the factors that influence listing prices has become increasingly important for hosts and travelers. The data consist of $1,863$ Airbnb listings from Quebec City, Canada, collected in June 2026 from the publicly available  platform~\cite{insideairbnb2026}. The response variable $Y$ is the log-transformed listing price, and the covariates include geographic location (\texttt{latitude} and \texttt{longitude}), listing characteristics (\texttt{minimum nights}, \texttt{room type}), host information (\texttt{calculated host listings count}), and booking activity measures (\texttt{number of reviews}, \texttt{reviews per month}, \texttt{number of reviews} in the last 12 months, and \texttt{availability} throughout the year). The categorical variable \texttt{room type} is embedded using one-hot encoding. For Airbnb listing price data, the GBM model uses the default setting in R. The neural network is a fully connected feed-forward model with hidden layers of 32 and 16 neurons, learning rate $10^{-3}$, weight decay $10^{-3}$, and batch size 64. From Table \ref{air_tb}, we see that GCM consistently achieves the lower averaged MSE across all different machine learning learners. Fig.\ref{air_stability} shows that GCM consistently identifying the significant variables in predicting listing price, such as \texttt{latitude}, \texttt{longitude}, and \texttt{minimum\_nights}. Across all learners, GCM generally tends to select more variables than LOCO and yields more stable selections.

\begin{figure}[t!]
    \centering     \includegraphics[width=0.9\textwidth]  {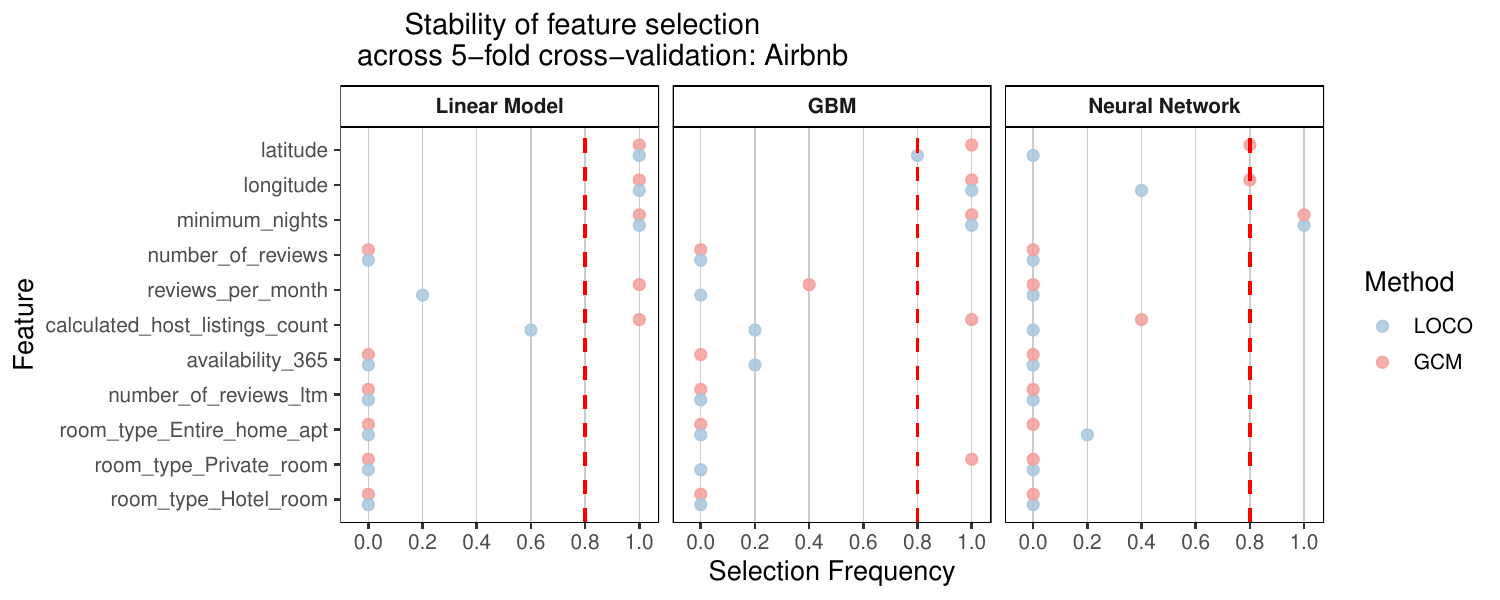}
    \caption{Selection rate of each feature selected by GCM and LOCO across 5-fold cross validation on Airbnb listing price data, using lm, GBM and NN as the base predictor respectively. Red dashed line: targeted selection frequency. }
    \label{air_stability}
\end{figure}
\begin{table}[t!]
    \centering
 \caption{Mean squared prediction error averaged across 5-fold cross-validation for GCM and LOCO 
on Airbnb listing price data, using lm, GBM, NN respectively. }
\begin{tabular}{l|ccc}
\hline & \multicolumn{3}{|c}{ Models: } \\
Method & lm & GBM & NN  \\
\hline GCM & 0.3338 $\pm$ 0.0093 & 0.2507 $\pm$ 0.0134 & 0.2892 $\pm$ 0.0113 \\
LOCO & 0.3366 $\pm$ 0.0105 & 0.2960 $\pm$ 0.0103 & 0.3227 $\pm$ 0.0272\\
\hline
\end{tabular}
\label{air_tb}
\end{table}

\subsection{Social media addiction dataset}
Whether social media use enhances or harms productivity has become an increasingly important question as digital platforms play a central role in everyday life. The dataset is publicly available on Kaggle~\cite{asif_zaman_2026} and contains $4,999$ observations after removing missing values. The response variable is \texttt{productivity score}, and the covariates include \texttt{age}, \texttt{daily screen time}, \texttt{social media hours}, \texttt{study hours}, \texttt{sleep hours}, \texttt{notifications per day}, \texttt{focus score}, and \texttt{addiction level}. For social media addiction data, the GBM model uses 500 trees and depth 4 in R. The neural network is a fully connected feed-forward model with hidden layers of 32 and 16 neurons, learning rate $10^{-3}$, weight decay $10^{-3}$, and batch size 128. Fig.\ref{social_stability} shows that both GCM and LOCO select \texttt{social media hours}, \texttt{study hours}, \texttt{sleep hours}, and the number of \texttt{notifications per day} consistently across learners. Compared with LOCO, GCM exhibits more stable selection, with same or higher frequencies for important variables and lower frequencies for insignificant variables. Table \ref{social_tb} shows that the two methods have identical prediction error under the linear model, while GCM achieves slightly lower mean squared prediction error than LOCO under GBM and neural network. 

\begin{figure}[tbhp]
    \centering     \includegraphics[width=0.9\textwidth]  {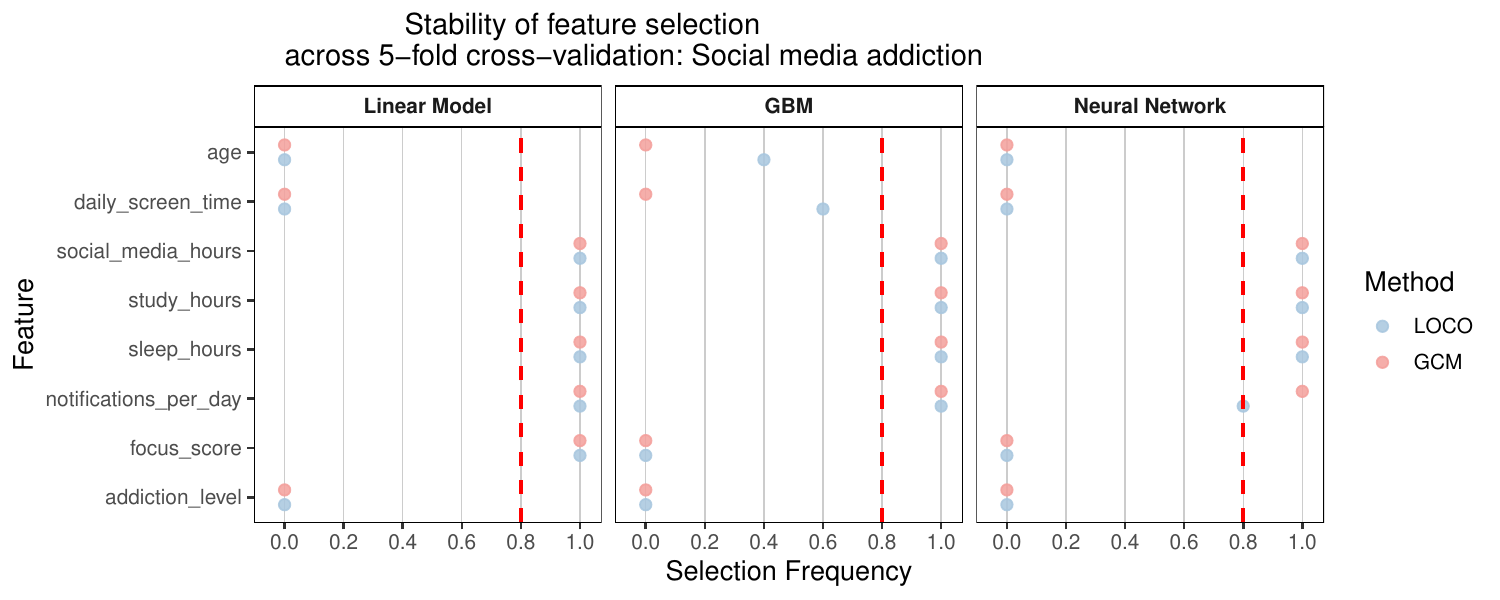}
    \caption{Selection rate of each feature selected by GCM and LOCO across 5-fold cross validation on impact of social media addiction on productivity data, using lm, GBM and NN as the base predictor respectively. Red dashed line: targeted selection frequency. }
    \label{social_stability}
\end{figure}
\begin{table}[t!]
    \centering
 \caption{Mean squared prediction error averaged across 5-fold cross-validation for GCM and LOCO 
on impact of social media addiction on productivity data, using lm, GBM, NN respectively. }
\begin{tabular}{l|ccc}
\hline & \multicolumn{3}{|c}{ Models: } \\
Method & lm & GBM & NN  \\
\hline GCM & 95.95 $\pm$ 2.56 & 91.23 $\pm$ 2.08 & 84.70 $\pm$ 1.95 \\
LOCO & 95.95 $\pm$ 2.56 & 91.85 $\pm$ 2.02  &  85.61 $\pm$ 2.31 \\
\hline
\end{tabular}
\label{social_tb}
\end{table}

\section{Conclusion}\label{sec:conclusion}
A key result of this paper is that GCM demonstrates greater efficiency in variable selection compared to LOCO under suitable regular conditions for linear, additive and single index models. Application to real-data variable selection further confirms that GCM selects models with lower prediction MSE than LOCO and is more stable. Future work could develop a novel GCM-based test statistic that improves computational efficiency or strikes a compromise with Type II error, advancing hypothesis testing methodologies.

\section*{Supplementary materials}
The online Supplementary Material contains an extra efficiency comparison example for additive model (Section \ref{app:s1}). Mean and variance for test statistics under different models (Section \ref{app_A}), proofs of theorems (Section \ref{app:proof}), and additional simulation and data analysis results (Section \ref{app_c}).
\par

\bibliographystyle{plain}
\bibliography{ref}
\newpage
\begin{appendices}
\section{Additional additive model example}\label{app:s1}

\begin{example}\label{ex3}
    Let $Y = cos(X_1) + sin(X_2) + \epsilon$, where $\epsilon \sim N(0,\sigma^2)$. We have $X_j \sim N(0,1)$, $X_j$'s are independent, and we have  $\tilde{X_j}  = X_j - E(X_j|X_{-j}) = X_j$.

    For $cos(X_1)$, we have $\tilde{f}_1(X_1) = cos(X_1) - E(X_1|X_{2})$, and then the  $\mbox{Corr}^2(\tilde{X}_1, \tilde{f}_{1}(X_1)) = 0$ due to the independence assumption and the symmetry of cosine function. Thus, the even function with symmetric distribution of $X$ violates the condition (A) in Theorem 4, and LOCO estimator is more efficient than GCM estimator.
    
    For $sin(X_2)$, we have $\tilde{f}_2(X_2) = sin(X_2)$. By Taylor series expansion, let's approximate $sin(X_2) \approx X_2 -\frac{X_2^3}{3!}+\frac{X_2^5}{5!}$. The numerator in RHS of the inequality in condition (A) is bounded by
    \vspace{0.01 cm}
\begin{equation}\label{three}
    \begin{split}
       E(\tilde{X_2}^2 \tilde{f_{2}}^2 )\frac{E(\tilde{f_{2}}^2)}{E(\tilde{X_2}^2)}  + \sigma^2 E(\tilde{f_2}^2) 
        &\leq E[X_2^{2} (3X_2^{p_j})^2]E[(3X_2)^{2*p_j}]+\sigma^2E[(3X_2^{p_j})^{2}] \\
        &\approx 81 E(X_2^{2+2p_j})E(X_2^{2p_j})+9\sigma^2E(X_2^{2p_j}).
    \end{split}
\end{equation}
Let $p_j = 5$ based on the approximation chosen above, and plug into Eq.(3.1) and (3.2) derived from Example 3.2 in the main context. We see that the upper bound (\ref{three}) of RHS of the numerator in condition (A) is less than  the lower bound of  $\mbox{Corr}^2$ times LHS of denominator in condition (A) (ie: $E(X_j^{4p_j})\frac{E^2(X_j^{p_j+1})}{E(X_j^{2p_j})} + 4\sigma^2E^2(X_j^{p_j+1})$). Thus, GCM estimator is more efficient than LOCO estimator up to the Taylor approximation of the sine function.

\end{example}
\vspace{-0.3 cm}

\section{Mean and variance for test statistics under different models}\label{app_A}

In the following sections, we compute all the expectation and variance of test statistics for different models. We will focus on the expression of population parameters since the expectation and variance of the plug-in estimator can directly be derived, i.e., $E(\hat{\psi}_{0,j}^{(\cdot)}) =E(\psi_{0,j}^{(\cdot)}) $ and $Var(\hat{\psi}_{0,j}^{(\cdot)}) =\frac{1}{n}Var(\psi_{0,j}^{(\cdot)}) $.
\subsection{Linear Model} 
Recall linear model setup: $Y = X_{-j}\beta_{-j}+\mathnormal{\beta}_j X_j+\epsilon$ where $\epsilon$ are i.i.d  mean zero with finite variance $\sigma^2$, and $\epsilon \perp  X$.
\paragraph{GCM test statistics}
For GCM hypothesis testing of each feature $j$, we have the product of residuals from regression as: 
\[X_j=h_{0,j}(X_{-j}) + \xi_{-j};\;\; Y= g_{0,j}(X_{-j}) + \epsilon_{-j}\] 
where $h_{0,j}(X_{-j}) = E(X_j|X_{-j}), g_{0,j}(X_{-j}) = E(Y|X_{-j})= X_{-j}\beta_{-j}+\beta_j E(X_j|X_{-j})$, and $\tilde{X_j} : = X_j - E(X_j|X_{-j})$. 
\begin{equation*}
    \begin{split}
        E(\psi_{0,j}^{(gcm)}) &= E[(X_j - E(X_j|X_{-j}))(Y-E(Y|X_{-j}))]\\
        &= E[(X_j - E(X_j|X_{-j}))(X_{-j}\beta_{-j}+\beta_jX_j+\epsilon- X_{-j}\beta_{-j} - \beta_j E(X_j|X_{-j}))]\\
        &=E\big\{[X_j - E(X_j|X_{-j})][\beta_j(X_j - E(X_j|X_{-j}))+\epsilon]\big\}\\
        &=\beta_j E(\tilde{X}_j^2);
    \end{split}
\end{equation*}
\begin{equation*}
    \begin{split}
        Var(\psi_{0,j}^{(gcm)}) &= E(\psi_{0,j}^{{(gcm)}^2}) - (E(\psi_{0,j}^{(gcm)}))^2\\
        &= E\big\{[X_j - E(X_j|X_{-j})]^2[\beta_j(X_j -E (X_j|X_{-j}))+\epsilon]^2\big\} -\beta_j^2Var^2(X_j|X_{-j}) \\
        &= \beta_j^2Var(\tilde{X_j} ^2) +\sigma^2 E(\tilde{X}_j^2).
    \end{split}
\end{equation*}
\paragraph{LOCO test statistics}
In general,
\begin{equation*}
    \begin{split}
        E(\psi_{0,j}^{(loco)}) 
        &=E[Y-g_{0,j}(X_{-j})]^2-E[Y-f_0(X)]^2\\
        &= E[(f_0(X)+\epsilon-g_{0,j}(X_{-j}))^2-(f_0(X)+\epsilon-f_0(X))^2]\\
        &=E[(f_0(X)-g_{0,j}(X_{-j}))^2].
    \end{split}
\end{equation*}
Under the setting of multivariate linear regression model, we have 
\begin{equation*}
    \begin{split}
       & f_0(X) = E(Y|X) = X_{-j}\beta_{-j}+\beta_jX_j\\
        &g_{0,j}(X_{-j}) =E(Y|X_{-j}) = X_{-j}\beta_{-j}+\beta_j E(X_j|X_{-j})
    \end{split}
\end{equation*}
Then, $E[\psi_{0,j}^{(loco)}] = E[(f_0(X)-g_{0,j}(X_{-j}))^2]=\beta_j^2E(X_j - E(X_j|X_{-j}))^2 =\beta_j^2 E(\tilde{X}_j^2);$
\begin{equation*}
    \begin{split}
        Var(\psi_{0,j}^{(loco)})
        &= Var[(f_0(X)-g_{0,j}(X_{-j}))^2+2\epsilon(f_0(X)-g_{0,j}(X_{-j})))]\\
        &=E[(f_0(X)-g_{0,j}(X_{-j}))^2+2\epsilon(f_0(X)-g_{0,j}(X_{-j})))]^2-\beta_j^4 E^2(\tilde{X}_j^2)\\
        &=\beta_j^2\big\{\beta_j^2E[(X_j - E(X_j|X_{-j}))^4 ]+4\sigma^2Var(X_j|X_{-j})-\beta_j^2E^2(\tilde{X}_j^2)\big\}\\
        &=\beta_j^4 Var(\tilde{X_j}^2)+4\sigma^2\beta_j^2 E(\tilde{X}_j^2).
    \end{split}
\end{equation*}
\paragraph{Dropout test statistics}
For Dropout estimator, let's recall $X^{(j)}$ is the $j^{th}$ column replaced by its marginal mean $\mu_j$. In general,
\begin{equation*}
    \begin{split}
          E(\psi_{0,j}^{(dr)}) &= V(f_{0},P_{0,-j}) - V(f_0,P_0)\\
        &=E[Y-f_{0}(X^{(j)})]^2-E_0[Y-f_0(X)]^2\\
        &=E[(f_0(X)-f_{0}(X^{(j)}))^2+2\epsilon(f_0(X)-f_{0}(X^{(j)}))]\\
        & = E[(f_{0}(X)-f_{0}(X^{(j)}))^2];
    \end{split}
\end{equation*}
Under the setting of multivariate linear regression model, we have
\begin{equation*}
    \begin{split}
       & f_0(X) = E(Y|X) = X_{-j}\beta_{-j}+\beta_jX_j,\\
        &f_{0}(X^{(j)})  = X_{-j}\beta_{-j}+\beta_j \mu_j.
    \end{split}
\end{equation*}
Then, $E[\psi_{0,j}^{(dr)}] = E[(f_{0}(X)-f_{0}(X^{(j)}))^2]= \beta_j^2 E((X_j -\mu_j)^2) = \beta_j^2E(\hat{X}_j^2)$,  where $\hat{X}_j:=X_j - E(X_j)$;
\begin{equation*}
    \begin{split}
        Var(\psi_{0,j}^{(dr)})
        &= Var[(f_0(X)-f_{0}(X^{(j)}))^2+2\epsilon(f_0(X)-f_{0}(X^{(j)}))]\\
        &=E[(f_0(X)-f_{0}(X^{(j)}))^2+2\epsilon(f_0(X)-f_{0}(X^{(j)}))]^2-\beta_j^4 E^2(\hat{X}_j^2)\\
        &=\beta_j^2\big\{\beta_j^2E[(X_j - \mu_j)^4 ]+4\sigma^2E(\hat{X}_j^2)-\beta_j^2E^2(\hat{X}_j^2)\big\}\\
        &=\beta_j^4 Var(\hat{X_j}^2)+4\sigma^2\beta_j^2 E(\hat{X}_j^2).
    \end{split}
\end{equation*}
 
\subsection{Non-linear Additive Model}
In this section, we compute all the expressions of test statistics under non-linear regression system.
Recall the non-linear additive model $Y = f_0(X) + \epsilon = f_{0,-j}(X_{-j})+f_{0,j}( X_j)+\epsilon$ where $\epsilon$ are i.i.d  mean zero with finite variance $\sigma^2$, and $\epsilon \perp  X$. Let $h_{0,j}(X_{-j}) = E(X_j|X_{-j}), g_{0,j}(X_{-j}) = E(Y|X_{-j})$, and $\tilde{f}_{0,j}(X_j):=f_{0,j}(X_j) - E(f_{0,j}(X_j)|X_{-j})$.
\paragraph{GCM test statistics}\label{app_A_non_lin_gcm}
\begin{equation*}
    \begin{split}
        E(\psi_{0,j}^{(gcm)}) &= E[(X_j - E(X_j|X_{-j}))(Y-E(Y|X_{-j}))]\\
        &= E[(X_j - E(X_j|X_{-j}))(f_{0,-j}(X_{-j})+f_{0,j}( X_j)+\epsilon- f_{0,-j}(X_{-j})\\
        &\hspace{5cm}- E(f_{0,j}(X_j)|X_{-j}))]\\
        &=E\big\{[X_j - E(X_j|X_{-j})][f_{0,j}(X_j )- E(f_{0,j}(X_j)|X_{-j})+\epsilon]\big\}\\
        &=E(\tilde{X}_j\tilde{f}_{0,j}(X_j)) ;
    \end{split}
\end{equation*}
\begin{equation*}
    \begin{split}
        Var(\psi_{0,j}^{(gcm)}) &= E(\psi_{0,j}^{{(gcm)}^2}) - (E(\psi_{0,j}^{(gcm)}))^2\\
        &=E\big\{[X_j - E(X_j|X_{-j})]^2[f_{0,j}(X_j )- E(f_{0,j}(X_j)|X_{-j})+\epsilon]^2\big\} \\
        & \hspace{2cm}-\bigg\{ E\big\{[X_j -E(X_j|X_{-j})][f_{0,j}(X_j )- E(f_{0,j}(X_j)|X_{-j})]\big\}\bigg\}^2\\
        &=Var(\tilde{X}_j\tilde{f}_{0,j}(X_j)) + \sigma^2 E(\tilde{X}_j^2).
    \end{split}
\end{equation*}
\paragraph{LOCO test statistics }\label{app_A_non_lin_loco}
Let $f_{0,j}(X) = E(Y|X)=f_{0,-j}(X_{-j})+f_{0,j}( X_j)$, and $ g_{0,j}(X_{-j}) = E(Y|X_{-j})=f_{0,-j}(X_{-j})+E(f_{0,j}( X_j)|X_{-j})$. Then, $E(\psi_{0,j}^{(loco)})  = E(f_{0,j}(X_j) - E(f_{0,j}(X_j)|X_{-j}))^2 = E(\tilde{f}_{0,j}^2(X_j))$. By the same computation process in linear system, we have $Var(\psi_{0,j}^{(loco)})  =Var[(f_0(X)-g_{0,j}(X_{-j}))^2+2\epsilon(f_0(X)-g_{0,j}(X_{-j}))]=Var(\tilde{f}_{0,j}^2(X_j)) + 4\sigma^2 E(\tilde{f}_{0,j}^2(X_j))$.

\paragraph{Dropout test statistics }

\begin{equation*}
    \begin{split}
        E(\psi_{0,j}^{(dr)}) &=  E[(f_{0}(X)-f_{0}(X^{(j)}))^2]\\
        &= E[(f_{0,j}(X_j)-f_{0,j}(\mu_j))^2]\\
        & = E[(f_{0,j}(X_j)-f_{0,j}(\mu_j) -E(f_{0,j}(X_j)) + E(f_{0,j}(X_j))^2] \\
        &= Var(f_{0,j}(X_j)) + (E(f_{0,j}(X_j)) - f_{0,j}(\mu_j))^2;
    \end{split}
\end{equation*}

\begin{equation*}
    \begin{split}
        Var(\psi_{0,j}^{(dr)}) &= Var[(Y-f_{0}(X^{(j)}))^2-(Y-f_0(X))^2]\\
        &= Var[(f_0(X)-f_{0}(X^{(j)}))^2+2\epsilon(f_0(X)-f_{0}(X^{(j)}))]\\ 
        &=E[(f_{0,j}(X_j)-f_{0,j}(\mu_j))^4] + 4 \sigma^2 [Var(f_{0,j}(X_j)) + (E(f_{0,j}(X_j)) - f_{0,j}(\mu_j))^2] \\
        &\hspace{3cm}-[Var(f_{0,j}(X_j)) + (E(f_{0,j}(X_j)) - f_{0,j}(\mu_j))^2]^2. 
        \end{split}
\end{equation*}
Expansion of $4^{th}$ moment:
\begin{equation*}
    \begin{split}
        E[(f_{0,j}(X_j)-f_{0,j}(\mu_j))^4] &= E\{[f_{0,j}(X_j)-E(f_{0,j}(X_j)) +E(f_{0,j}(X_j)) - f_{0,j}(\mu_j)]^4\}\\
        &= E[(f_{0,j}(X_j) - E(f_{0,j}(X_j)))^4]\\
        &+6Var(f_{0,j}(X_j))(E(f_{0,j}(X_j)) - f_{0,j}(\mu_j))^2+(E(f_{0,j}(X_j)) - f_{0,j}(\mu_j))^4.
    \end{split}
\end{equation*}
Thus,
\begin{equation*}
    \begin{split}
Var(\psi_{0,j}^{(dr)}) 
&= E[(f_{0,j}(X_j) - E(f_{0,j}(X_j)))^4]- Var^2(f_{0,j}(X_j))+4\sigma^2Var(f_{0,j}(X_j)) \\
&+4(E(f_{0,j}(X_j)) - f_{0,j}(\mu_j))^2(Var(f_{0,j}(X_j))+\sigma^2)\\
&=Var(\hat{f}_{0,j}^2(X_j)) + 4\sigma^2E(\hat{f}^2_{0,j}(X_j))+4(E(f_{0,j}(X_j)) - f_{0,j}(\mu_j))^2(Var(f_{0,j}(X_j))+\sigma^2)
    \end{split}
\end{equation*}
where $\hat{f}_{0,j}(X_j):=f_j(X_j) - E(f_j(X_j))$.

\subsection{Single Index Model}\label{app_A_SIM}
\vspace{-1 em}
In this section, we will compute all the expressions of test statistics in single index model which is a semi-parametric extension of linear model. Recall the single index model $Y = \eta(X^T\beta) + \epsilon$ where $\beta$ is unknown parameter vector, $\eta$ is an unknown link function, and $\epsilon$ are i.i.d errors with zero mean and finite variance $\sigma^2$, and $\epsilon \perp X$.

Suppose under the conditions in Theorem 5 it is sufficient to approximate $\eta$ via linear Taylor series expansion and the remainder term $\Delta$ involving higher order terms in the expansion is negligible. That is to say that $\Delta, E(\Delta|X_{-j}) = o(1)$, and we will prove it later in Appendix \ref{app_B_SIM}. Thus, the test statistics derivation procedure is similar to linear model section.
Specifically, let's recall the Taylor expansion of $\eta(X^T\beta)$ around \( E(X^T \beta | X_{-j}) \), and notation $\eta' =\eta'(E(X^T \beta | X_{-j})).$

\begin{equation*}
    \begin{split}
      f_0(X) =E(Y|X)&= \eta(E(X^T \beta | X_{-j})) + \eta' \cdot (X^T \beta - E(X^T \beta | X_{-j}))+\Delta\\
      & = \eta(E(X^T \beta | X_{-j})) + \eta' \cdot \beta_j (X_j -   E(X_j  | X_{-j}))+\Delta\\
      &\text{where $\Delta = \sum_{s=2}^{\infty}\frac{1}{s!}\eta^{(s)}(E(X^T\beta|X_{-j}))(X^T \beta - E(X^T \beta | X_{-j}))^s$.}
    \end{split}
\end{equation*}

\paragraph{GCM test statistics}
\begin{equation*}
    \begin{split}
        E (\psi_{0,j}^{(gcm)}) &= E [(X_j -   E (X_j  | X_{-j}))(Y - E (Y|X_{-j}))]\\
        &= E [(X_j -   E (X_j  | X_{-j}))(\eta(X^T\beta)+\epsilon - \eta(E (X^T \beta | X_{-j})))- E (\Delta|X_{-j})] \\
        &= E [(X_j -   E (X_j  | X_{-j}))(\eta' \beta_j (X_j -   E (X_j  | X_{-j}))+\Delta - E (\Delta|X_{-j}) ]\\
        & = \beta_j E (\tilde{X}_j^2)\eta' + E [(X_j -   E (X_j  | X_{-j}))o(1)]\\
        &=\beta_j E (\tilde{X}_j^2)\eta'+o(1);
    \end{split}
\end{equation*}

\begin{equation*}
    \begin{split}
        Var(\psi_{0,j}^{(gcm)})  &= E(\psi_{0,j}^{{(gcm)}^2}) - (E(\psi_{0,j}^{(gcm)}))^2\\
        &=  E[\tilde{X}_j^2 (\eta(X^T\beta)+\epsilon - \eta( E(X^T \beta | X_{-j})- E(\Delta|X_{-j}]))^2]  - (\beta_j E(\tilde{X}_j^2)\eta'+o(1))^2\\
        &= E[\tilde{X}_j^2(\eta' \beta_j \tilde{X}_j+\epsilon+o(1))^2] - \beta_j^2  E^2(\tilde{X}_j^2)\eta'^2+o(1)\\
        &=\eta'^2 \beta_j^2  E(\tilde{X}_j^4) + [\sigma ^2+o(1)] E(\tilde{X}_j^2)+2o(1)\eta'\beta_j E(\tilde{X}_j^3)- \beta_j^2  E^2(\tilde{X}_j^2)\eta'^2+o(1)\\
        &=\eta'^2 \beta_j^2Var(\tilde{X}_j^2)+ \sigma ^2 E(\tilde{X}_j^2 ) + o(1).
    \end{split}
\end{equation*}

\paragraph{LOCO test statistics}
\begin{equation*}
    \begin{split}
        E[\psi_{0,j}^{(loco)}]& = E[(E(Y|X) - E(Y|X_{-j}))^2] \\
        &= E[(\eta' \beta_j (X_j -   E(X_j  | X_{-j})) +\Delta - E(\Delta|X_{-j}))^2 ]\\
        & = \beta_j^2 E(\tilde{X}_j^2)\eta'^2+o(1);
    \end{split}
\end{equation*}

\begin{equation*}
    \begin{split}
        Var[\psi_{0,j}^{(loco)}]& = Var[(E(Y|X)-E(Y|X_{-j}))^2+2\epsilon(E(Y|X)-E(Y|X_{-j})))]\\
        &= \beta_j^4 E(\tilde{X}_j^4)\eta'^4+4\sigma^2\beta_j^2E(\tilde{X}_j^2)\eta'^2-\beta_j^4 E^2(\tilde{X}_j^2)\eta'^4+o(1)\\
        &=\beta_j^4 Var(\tilde{X}_j^2)\eta'^4+4\sigma^2\beta_j^2E(\tilde{X}_j^2)\eta'^2+o(1).\\
    \end{split}
\end{equation*}

\paragraph{Dropout test statistics}
Under the single index model, let's make estimation of $f_0(X^{(j)})$ also via first order Taylor series expansion around $E(X^T\beta|X_{-j})$:
$f_0(X^{(j)}) = \eta(\beta_j\mu_j + X_{-j}^T \beta_{-j}) 
       = \eta(E(X^T \beta | X_{-j})) + \eta' \cdot \beta_j (  \mu_j - E(X_j|X_{-j}))+\Delta$,
\begin{equation*}
    \begin{split}
    E[\psi_{0,j}^{(dr)}] &= E[(f_{0}(X)-f_{0}(X^{(j)}))^2]\\
    &= \beta_j^2 \eta'^2E((X_j -\mu_j)^2)+o(1)\\
    &= \beta_j^2\eta'^2E(\hat{X}_j^2)+o(1);\\
        Var(\psi_{0,j}^{(dr)})
        &= Var[(f_0(X)-f_{0}(X^{(j)}))^2+2\epsilon(f_0(X)-f_{0}(X^{(j)}))]\\
        &=\beta_j^2\eta'^2\big\{\beta_j^2\eta'^2E[(X_j - \mu_j)^4 ]+4\sigma^2E(\hat{X}_j^2)-\beta_j^2\eta'^2E^2(\hat{X}_j^2)\big\}+o(1)\\
        &=\beta_j^4 Var(\hat{X_j}^2)\eta'^4+4\sigma^2\beta_j^2 E(\hat{X}_j^2)\eta'^2+o(1).
    \end{split}
\end{equation*}

\section{Proofs of Theorems}\label{app:proof}
\subsection{Proof of Theorem 3}\label{app_B_lin}
Since the estimators of GCM and LOCO have different means: one is the product of residuals between models, and the other is the difference of MSE between models. Thus, we use the so-called \emph{relative variability} $\sigma/\mu$ to compare the asymptotic relative efficiency.

\begin{equation*}
    \begin{split}      e_{\hat{\psi}_{0,j}^{(gcm)},\hat{\psi}^{(loco)}_{0,j}} &= \bigg(\frac{sd({\hat{\psi}^{(loco)}_{0,j}}) / E(\hat{\psi}_{0,j}^{(loco)})}{sd({\hat{\psi}^{(gcm)}_{0,j}}) / E(\hat{\psi}_{0,j}^{(gcm)})}\bigg)^2\\
    &=\bigg(\frac{sd(\hat{\psi}^{(loco)}_{0,j})}{\beta_j sd(\hat{\psi}_{0,j}^{(gcm)})}\bigg)^2\\
        &=\frac{\beta_j^2\big\{\beta_j^2E[(X_j - E(X_j|X_{-j}))^4 ]+4\sigma^2Var(X_j|X_{-j})-\beta_j^2Var^2(X_j|X_{-j})\big\}}{\beta_j^2\{\beta_j^2E[X_j - E(X_j|X_{-j})]^4+\sigma^2Var(X_j|X_{-j})-\beta_j^2Var^2(X_j|X_{-j})\}}\\
        &=\frac{\beta_j^4 Var(\tilde{X_j}^2)+4\sigma^2\beta_j^2 E(\tilde{X}_j^2)}{\beta_j^4Var(\tilde{X_j} ^2) +\sigma^2 \beta_j^2E(\tilde{X}_j^2)}>1.\\
    \end{split}
\end{equation*}

\subsection{Proof of Theorem 4}
\begin{equation*}
\begin{split}               e_{\hat{\psi}_{0,j}^{(gcm)},\hat{\psi}^{(loco)}_{0,j}} &= \bigg(\frac{sd({\hat{\psi}^{(loco)}_{0,j}}) / E(\hat{\psi}_{0,j}^{(loco)})}{sd({\hat{\psi}^{(gcm)}_{0,j}}) / E(\hat{\psi}_{0,j}^{(gcm)})}\bigg)^2\\
       &=\bigg(\frac{sd(\hat{\psi}^{(loco)}_{0,j})}{\frac{E[\tilde{f}_{0,j}(X_j)^2]}{E[\tilde{X}_j\tilde{f}_{0,j}(X_j)]}sd(\hat{\psi}_{0,j}^{(gcm)})}\bigg)^2 \text{, replace $\tilde{f}_{0,j}(X_j)$ with notation $\tilde{f}_{0,j}$}\\
    &=\frac{Var(\hat{\psi}^{(loco)}_{0,j})}{\frac{1}{Corr^2(\tilde{X}_j, \tilde{f}_{0,j})}\frac{E[\tilde{f}_{0,j}^2]}{E[\tilde{X}_j^2]}Var(\hat{\psi}_{0,j}^{(gcm)})} \text{, replace $Corr^2(\tilde{X}_j, \tilde{f}_{0,j})$ with notation $\rho^2_{\tilde{X}_j, \tilde{f}_{0,j}}$}\\
       & = \frac{E(\tilde{f}_{0,j}^4)+4\sigma^2 E(\tilde{f}_{0,j}^2)-E^2(\tilde{f}_{0,j}^2)}{\big\{E[\tilde{X}_j^2 \tilde{f}_{0,j}^2 ]  + \sigma^2 E(\tilde{X}_j^2)-E^2(\tilde{X_j} \tilde{f}_{0,j})\big\}\frac{1}{\rho_{\tilde{X}_j, \tilde{f}_{0,j}}^2}\frac{E(\tilde{f}_{0,j}^2)}{E(\tilde{X_j}^2)}}\\
       &= \frac{E(\tilde{f}_{0,j}^4)\rho_{\tilde{X}_j, \tilde{f}_{0,j}}^2+4\sigma^2 \rho_{\tilde{X}_j, \tilde{f}_{0,j}}^2E(\tilde{f}_{0,j}^2)-\frac{E^2(\tilde{X}_j\tilde{f}_{0,j})}{E(\tilde{X}_j^2)}E^2(\tilde{f}_{0,j}^2)}{E[\tilde{X}_j^2 \tilde{f}_{0,j}^2 ]\frac{E(\tilde{f}_{0,j}^2)}{E(\tilde{X}_j^2)}  + \sigma^2 E(\tilde{f}_{0,j}^2)-\frac{E^2(\tilde{X}_j\tilde{f}_{0,j})}{E(\tilde{X}_j^2)}E^2(\tilde{f}_{0,j}^2)} >1\text{, if assumption (A) holds.}\\
\end{split}
\end{equation*}

\subsection{Proof of Theorem 5}\label{app_B_SIM}
    In this section, we first prove that the remainder term $\Delta$ in Taylor series expansion is negligible, so it is sufficient to apply linear approximation. Then we prove that the GCM estimator is asymptotically more efficient than LOCO. 

By assumption (B1) that $\eta^{(s)} \leq O(a^{-X^T\beta}) $ ,$a>1$ for all $s\geq 2$, assumption (B2) that $X$ has sub-Gaussian distribution with variance proxy $\sigma^2$, and assumption (B3) which makes $\beta_j$ to be finite, we can bound the higher order derivatives and polynomial terms in $\Delta$. For any \(\delta > 0\), with probability at least \(1 - \delta\):
\begin{equation*}
    \begin{split}
        \Delta &= \sum_{s=2}^{\infty}\frac{1}{s!}g^{(s)}(E(X^T\beta|X_{-j}))(X^T \beta - E(X^T \beta | X_{-j}))^s\\
        &\leq  \sum_{s=2}^{\infty}\frac{O(a^{-E(X^T\beta|X_{-j})}) \beta_j}{s!}(X_j - E(X_j  | X_{-j}))^s\\
        &\leq  \sum_{s=2}^{\infty}\frac{ \beta_j}{C' a^{E(X^T\beta|X_{-j})}s!}\sigma^s \left(\sqrt{2 \log\left(\frac{2}{\delta}\right)}\right)^s \text{, for some constant $C'$}\\
        &\leq \frac{ \beta_j}{C' a^{E(X^T\beta|X_{-j})}} e^{\sigma \sqrt{2\log (\frac{2}{\delta})}}\\
        &= o(1) 
    \end{split}
\end{equation*}
The last inequality holds due to the exponential series that for any $s\geq 2, \sum_{s=2}^{\infty} \frac{t^{s}}{s!}
= e^{t} - 1 - t
\le e^{t}$. Finally, the remainder term vanishes if there exists a constant $C$, such that $E(X^T\beta|X_{-j}) \geq C\sqrt{\log (\frac{1}{\delta})}$, then $\Delta \leq \epsilon$ for any $\epsilon >0$. Besides, we also have that
\begin{equation*}
    \begin{split}
        E(\Delta|X_{-j}) &= E[\sum_{s=2}^{\infty}\frac{1}{s!}g^{(s)}(E(X^T\beta|X_{-j}))(X^T \beta - E(X^T \beta | X_{-j}))^s|X_{-j}]\\
        &\leq  \sum_{s=2}^{\infty}\frac{O(a^{-E(X^T\beta|X_{-j})}) \beta_j}{s!}E[(X_j - E(X_j  | X_{-j}))^s|X_{-j}]\\
        &\leq  \sum_{s=2}^{\infty}\frac{ \beta_j}{C' a^{E(X^T\beta|X_{-j})}s!} (\sigma^2)^{\frac{s}{2}}s\Gamma(s/2)\text{, for some constant $C'$}\\
        & \leq \sum_{s=2}^{\infty}\frac{ \beta_j}{C' a^{E(X^T\beta|X_{-j})}s!} \sigma^ss(s/2)^{s/2}\\
        &\leq \sum_{s=2}^{\infty}\frac{ \beta_j}{C' a^{E(X^T\beta|X_{-j})}s!} \sigma^s K^s\text{, for some constant $K \geq 3$}\\
        &\leq \frac{ \beta_j}{C' a^{E(X^T\beta|X_{-j})}} e^{\sigma \sqrt{K}}\\
        &= o(1)
    \end{split}
\end{equation*}
Similarly, $E(\Delta|X_{-j})$ is negligible if there exists a constant $C$, such that $E(X^T\beta|X_{-j}) \geq C\sqrt{K}$, then $E(\Delta|X_{-j}) <\epsilon$. Thus, as stated in assumption B4), if there exists a constant C, such that $E(X^T\beta|X_{-j}) \geq C \max\{\sqrt\frac{1}{\delta},\sqrt{K}\}$, where the upper bound $K\geq 3$, then $\Delta, E(\Delta|X_{-j}) = o(1)$.
Next, with the estimator expressions derived in Supplementary Material \ref{app_A_SIM}, the same relative efficiency comparison technique, and monotonicity of $\eta$ in assumption (B1), we have that
\begin{equation*}
    \begin{split}
e_{\hat{\psi}_{0,j}^{(gcm)},\hat{\psi}^{(loco)}_{0,j}} &= \bigg(\frac{sd({\hat{\psi}^{(loco)}_{0,j}}) / E(\hat{\psi}_{0,j}^{(loco)})}{sd({\hat{\psi}^{(gcm)}_{0,j}}) / E(\hat{\psi}_{0,j}^{(gcm)})}\bigg)^2\\
    &=\bigg(\frac{sd(\hat{\psi}^{(loco)}_{0,j})}{\beta_j \eta'sd(\hat{\psi}_{0,j}^{(gcm)})}\bigg)^2 \text{ , recall notation $\eta' =\eta'(E(X^T \beta | X_{-j}))$}\\
        &=\frac{\beta_j^2\eta'^2\big\{\beta_j^2 \eta'^2Var(\tilde{X}_j^2))+4\sigma^2E(\tilde{X}_j^2)\big\}+o(1)}{\beta_j^2\eta'^2\{\beta_j^2\eta'^2 Var(\tilde{X}_j^2)+ \sigma ^2 E(\tilde{X}_j^2)\}+o(1)}\\
        &= \frac{\beta_j^2\eta'^2\big\{\beta_j^2 \eta'^2Var(\tilde{X}_j^2))+4\sigma^2E(\tilde{X}_j^2)\big\}}{\beta_j^2\eta'^2\{\beta_j^2\eta'^2 Var(\tilde{X}_j^2)+ \sigma ^2 E(\tilde{X}_j^2)\}} \bigg(1+o(1)\bigg) \\
    \end{split}
\end{equation*}
Thus, if assumption (B1)-(B4) hold, then for any $\epsilon' >0, \delta>0$, with probability at least $1-\delta$, we have  $e_{\hat{\psi}_{0,j}^{(gcm)},\hat{\psi}^{(loco)}_{0,j}} >1-\epsilon'.$
\begin{remark}
\textit{Note that the expressions in GCM and LOCO test statistics involve the estimate of $\eta'$. If $\eta$ is fixed and known, then the result can be directly computed. If $\eta$ is fixed but unknown, we can employ the Nadaraya-Watson kernel-weighted average estimation method proposed by \cite{foster_variable_2013} (see details in section 2.2) where $\eta'$ is estimated from the model $\tilde{y}_i = \eta'(\beta^T\tilde{x}_i) + \tilde{\epsilon}_i$ with $\tilde{x}_i = x_i + \frac{x_{i+1} - x_i}{2}$ , $\tilde{y}_i = \frac{\eta(\beta_0 x_{i+1}) - \eta(\beta_0^Tx_i)}{\beta_0 x_{i+1} - \beta_0^Tx_i}$, and known $\beta_0^TX$ (in our case, it is $E(\beta^T X|X_{-j})$).}

\end{remark}

\subsection{Proof of Theorem 7}\label{app_B_dr_loco}
Under the regularity conditions of neural network required in Theorem 4.4 of \cite{gao_lazy_2022}, the asymptotic variance of the Lazy-VI estimator $\hat{\psi}_{0,j}^{(lazy)}$ is the same
as that of LOCO estimator $\hat{\psi}_{0,j}^{(loco)}$, which is $\frac{1}{n}Var(\epsilon^2 - \epsilon^2_{-j})$, see reformulated Theorem 6 in main paper.

Next, let compare the asymptotic variance relationship of LOCO and dropout estimators. Under assumptions in Theorem 3, for $j \in [p]$, we have
\[ \frac{Var(\hat{\psi}^{(dr)}_{0,j})}{Var(\hat{\psi}^{(loco)}_{0,j})}=\frac{\beta_j^4 Var(\hat{X_j}^2)+4\sigma^2\beta_j^2 E(\hat{X}_j^2)}{\beta_j^4 Var(\tilde{X_j}^2)+4\sigma^2\beta_j^2 E(\tilde{X}_j^2)}\geq1.\]
The inequality holds since the unconditional variance of the dropout estimator is greater than or equal to the conditional variance of the LOCO estimator. This follows from the principle that unconditional moments are greater than or equal to their corresponding conditional moments which is  implied by Jensen's inequality for convex functions, such as variance.

Under assumptions in Theorem 4 , for all $j \in [p]$, similarly we have
\[ \frac{Var(\hat{\psi}^{(dr)}_{0,j})}{Var(\hat{\psi}^{(loco)}_{0,j})}
        =\frac{Var(\hat{f}_{0,j}^2(X_j)) + 4\sigma^2E(\hat{f}^2_{0,j}(X_j))+4(E(f_{0,j}(X_j)) - f_{0,j}(\mu_j))^2(Var(f_{0,j}(X_j))+\sigma^2)}{Var(\tilde{f}_{0,j}^2(X_j)) + 4\sigma^2 Var(\tilde{f}_{0,j}(X_j))}\geq1.\]       
The inequality holds due to the extra non-negative term and the replacement of the unconditional moments for the variance of dropout estimator.

Finally, under assumptions in Theorem 5, using the same Taylor series expansion approach as shown in Supplementary Material \ref{app_B_SIM}, we have $\Delta, E(\Delta|X_{-j}) = o(1)$. Then, with  monotonicity of $\eta$ in assumption (B1), we have for all $j \in [p]$
\begin{equation*}
    \begin{split}
       \frac{Var(\hat{\psi}^{(dr)}_{0,j})}{Var(\hat{\psi}^{(loco)}_{0,j})}
        &=\frac{\beta_j^4 Var(\hat{X_j}^2)\eta'^4+4\sigma^2\beta_j^2 E(\hat{X}_j^2)\eta'^2+o(1)}{\beta_j^4 Var(\tilde{X}_j^2)\eta'^4+4\sigma^2\beta_j^2E(\tilde{X}_j^2)\eta'^2+o(1)}\\
        & = \frac{\beta_j^4 Var(\hat{X_j}^2)\eta'^4+4\sigma^2\beta_j^2 E(\hat{X}_j^2)\eta'^2}{\beta_j^4 Var(\tilde{X}_j^2)\eta'^4+4\sigma^2\beta_j^2E(\tilde{X}_j^2)\eta'^2}\bigg(1+o(1)\bigg)     
    \end{split}
\end{equation*}
If assumption (B1)-(B4) in Theorem 5 hold, then for any $\epsilon' >0, \delta>0$, with probability at least $1-\delta$, we have $e_{\hat{\psi}_{0,j}^{(loco)},\hat{\psi}^{(dr)}_{0,j}} >1-\epsilon'$.

\section{Additional experiment results}\label{app_c}
In this section, we explore more extra simulation and real data analysis to prove the consistency of our theoretical result that GCM continues to outperform LOCO.

\subsection{Additional simulation}
Besides making a comparison for different feature selection methods with model-agnostic algorithm, we initially use canonical implementations for each type of the model. Specifically, we employ linear regression for (a), generalized additive model for (b) and (c),  and generalized partially linear single index model for (d) (\texttt{stats}, \cite{bolar_stat_2019} ; \texttt{gam}, \cite{hastie_gam_2024}; \texttt{gplsim}, \cite{zu_gplsim_2023} package respectively in \texttt{R}). 
Fig.~\ref{grouped_gcm_loco_uni_plots} shows the average p-value for each of the $20$ features performed by GCM, LOCO, dropout and permutation variable selection. 

\begin{figure}[t!]
    \centering
        \includegraphics[width=0.5\textwidth]{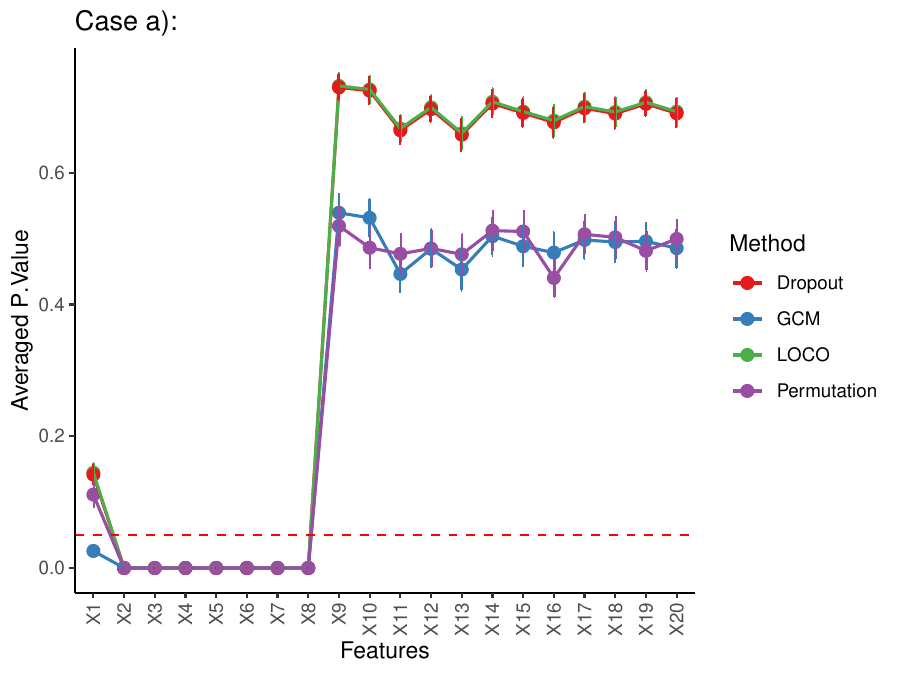}%
    \hfill
        \includegraphics[width=0.5\textwidth]{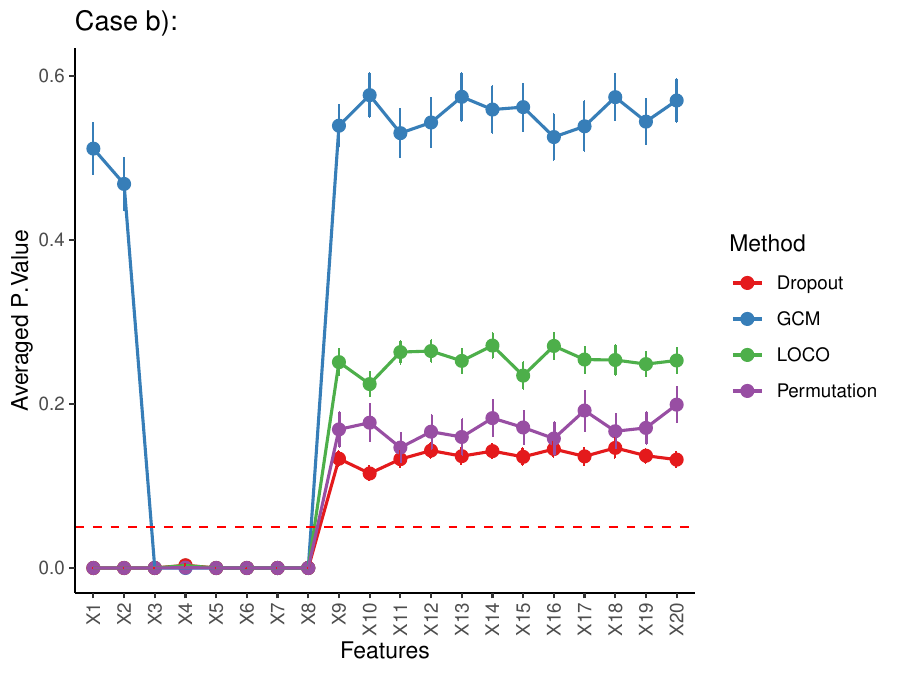}%

    \vspace{0.1cm}
        \includegraphics[width=0.5\textwidth]{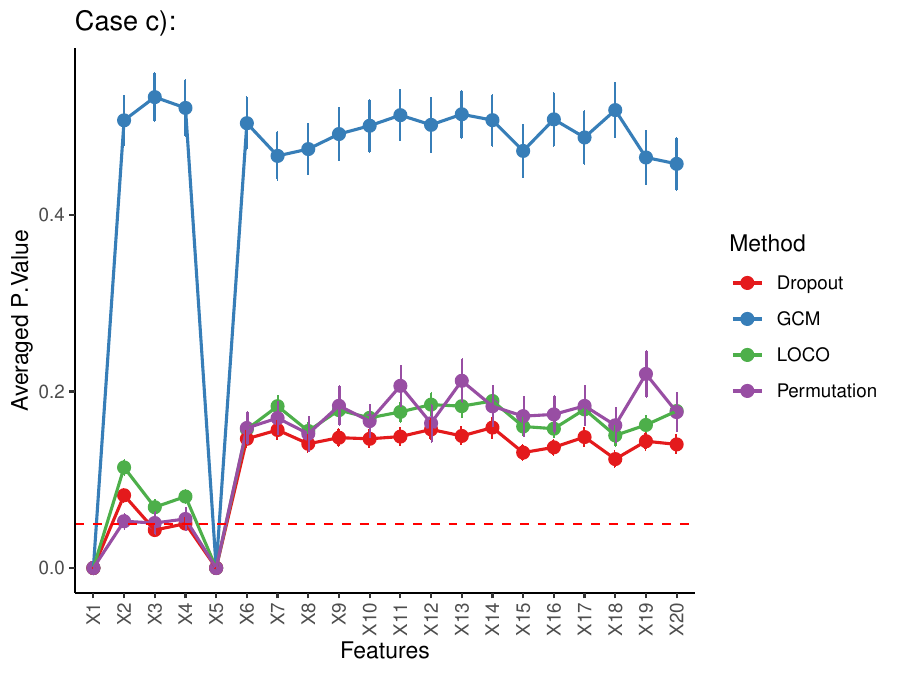}%
    \hfill
        \includegraphics[width=0.5\textwidth]{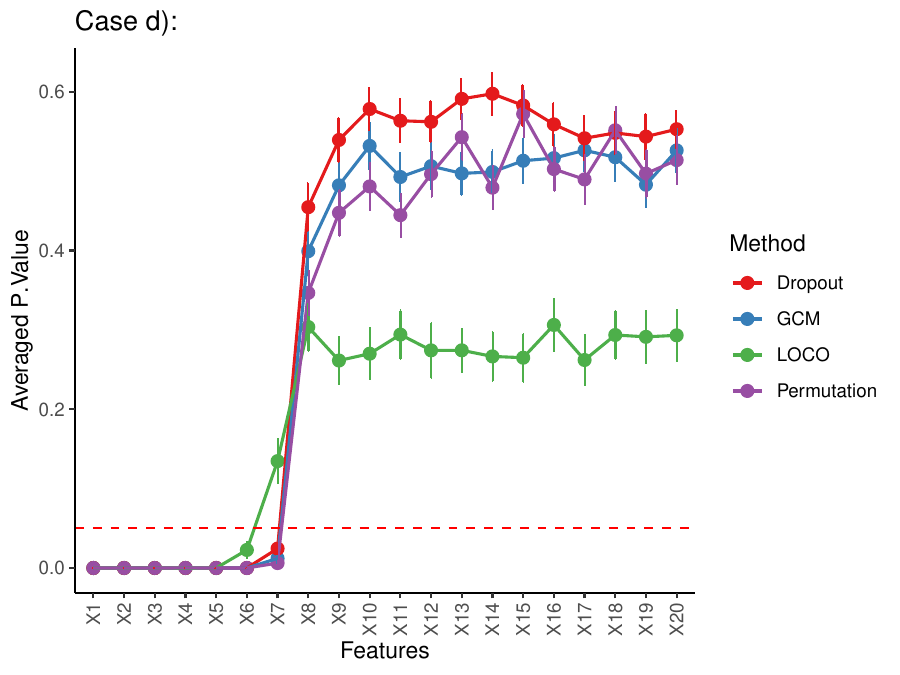}%
    
    \caption{Averaged p-value with standard deviation of each feature for model (a) - (d)  for GCM, LOCO, dropout and permutation univariate selection on dataset using linear regression for (a), generalized additive  model for (b) and (c), and generalized partially linear single index model for (d). Red dashed line: targeted type-I error rate ($\alpha = 0.05$). }
    \label{grouped_gcm_loco_uni_plots}
\end{figure}

In Fig.~\ref{grouped_gcm_loco_uni_plots}, plot (a) shows that all approaches work well in the linear model case. GCM demonstrates an advantage by accurately selecting variables with relatively 
small regression coefficients, highlighting its sensitivity to subtle but important relationships in the linear model. The performance of dropout, permutation and LOCO are almost the same due to the independence structure in model (a), and the moderate correlation between $X_3$ and $X_4$ does not make a large difference in p-values. For plot (b), as discussed in 
 Example 3.2 (main context) and Example \ref{ex3} (Supplementary material S1), GCM successfully captures all other features except the first two in the non-linear additive model (b) because the quadratic and cosine functions are even functions, which violates assumption (A). None of the methods work well when the true model involves interaction terms as shown in plot (c). Overall, all methods have similar performance when the standard algorithm is selected for the true model, and GCM performs slightly better than LOCO as seen in plot (a) and (d).
 \begin{figure}[t!]
        \includegraphics[width=0.5\textwidth]{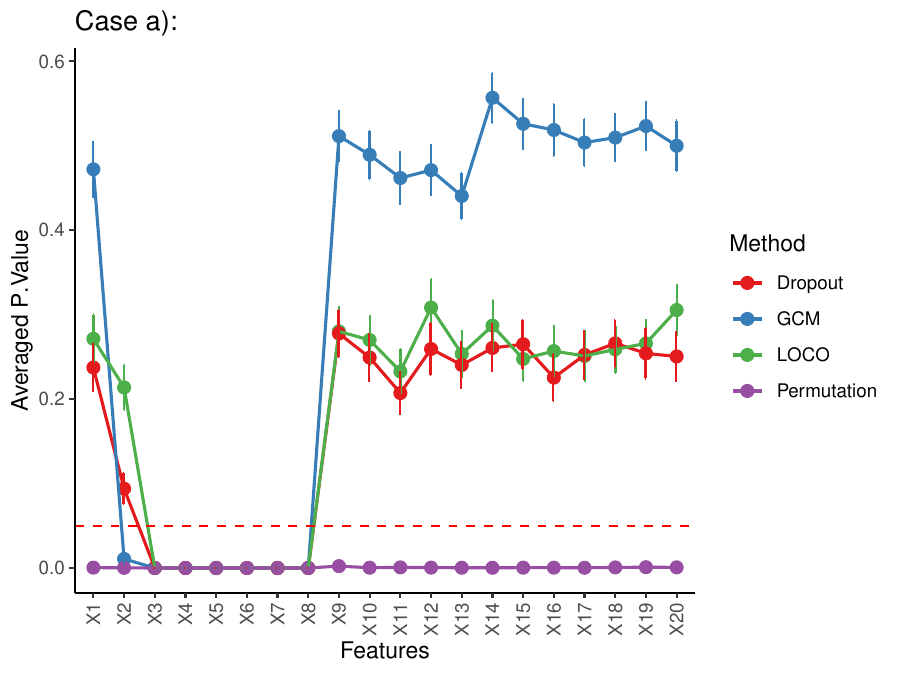}
    \hfill
        \includegraphics[width=0.5\textwidth]{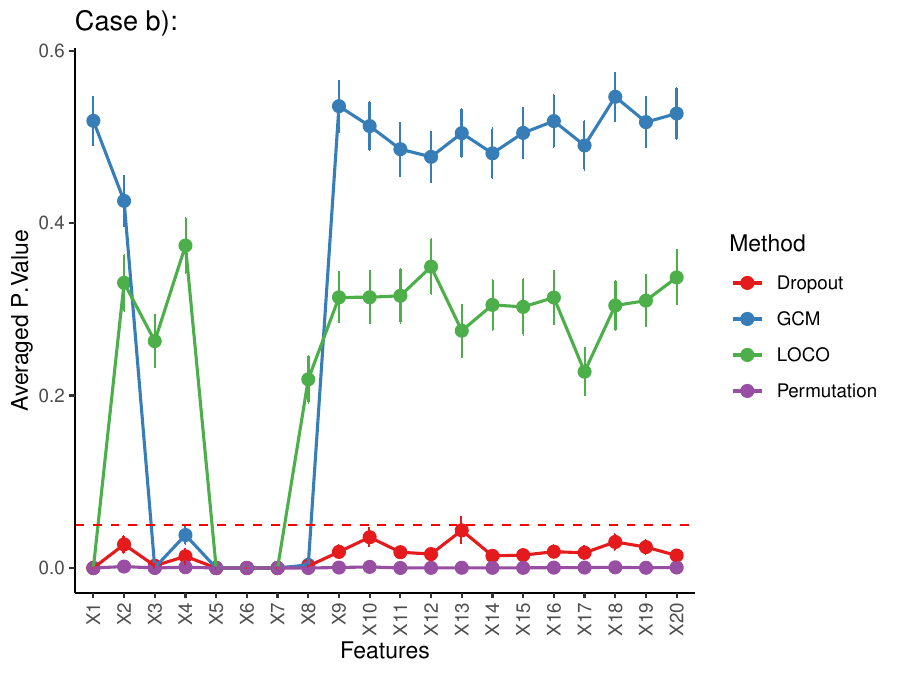}%

    \vspace{0.1cm}
        \includegraphics[width=0.5\textwidth]{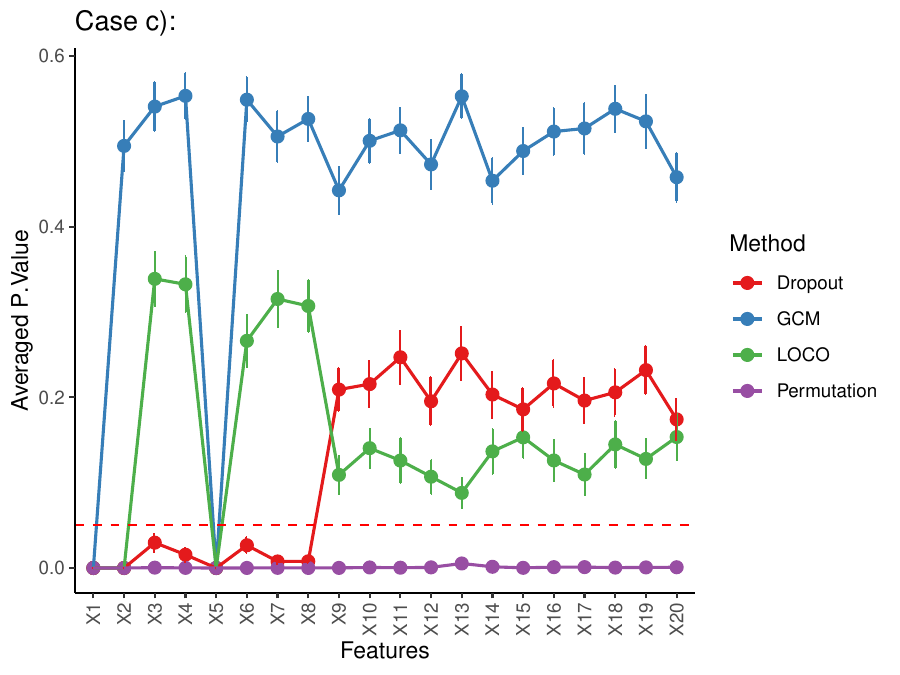}%
    \hfill
        \includegraphics[width=0.5\textwidth]{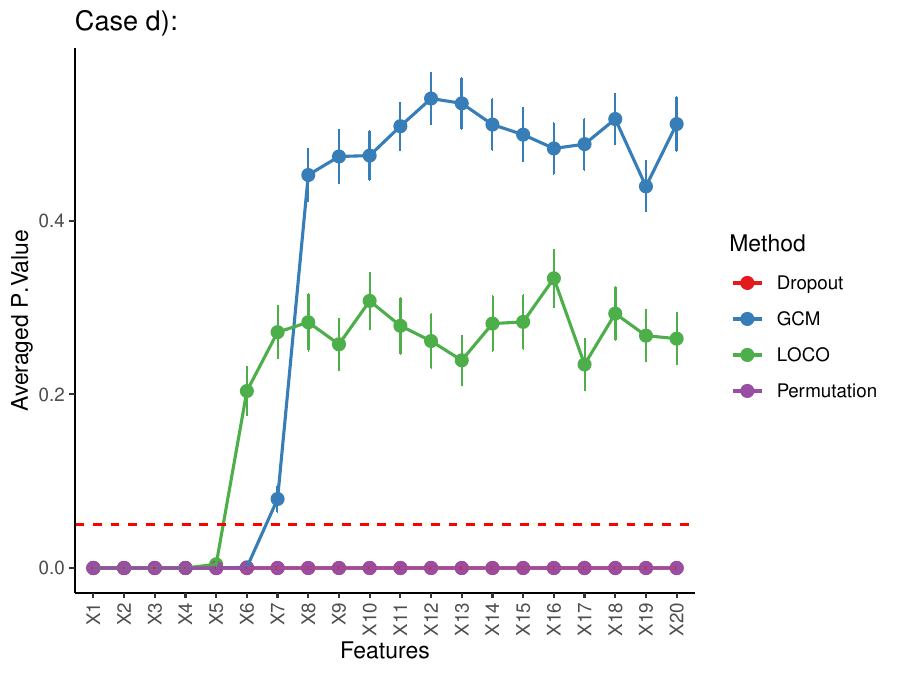}%
    
    \caption{Averaged p-value and standard deviation of each feature for model (a) - (d)  across 100 simulations for GCM, LOCO, dropout and permutation univariate selection on dataset with 1000 instances using kernel SVM. Red dashed line: targeted type-I error rate ($\alpha = 0.05$). }
    \label{gcm_loco_ksvm_plots}
\end{figure}

\begin{figure}[t!]
  \centering
  \includegraphics[width =\textwidth]{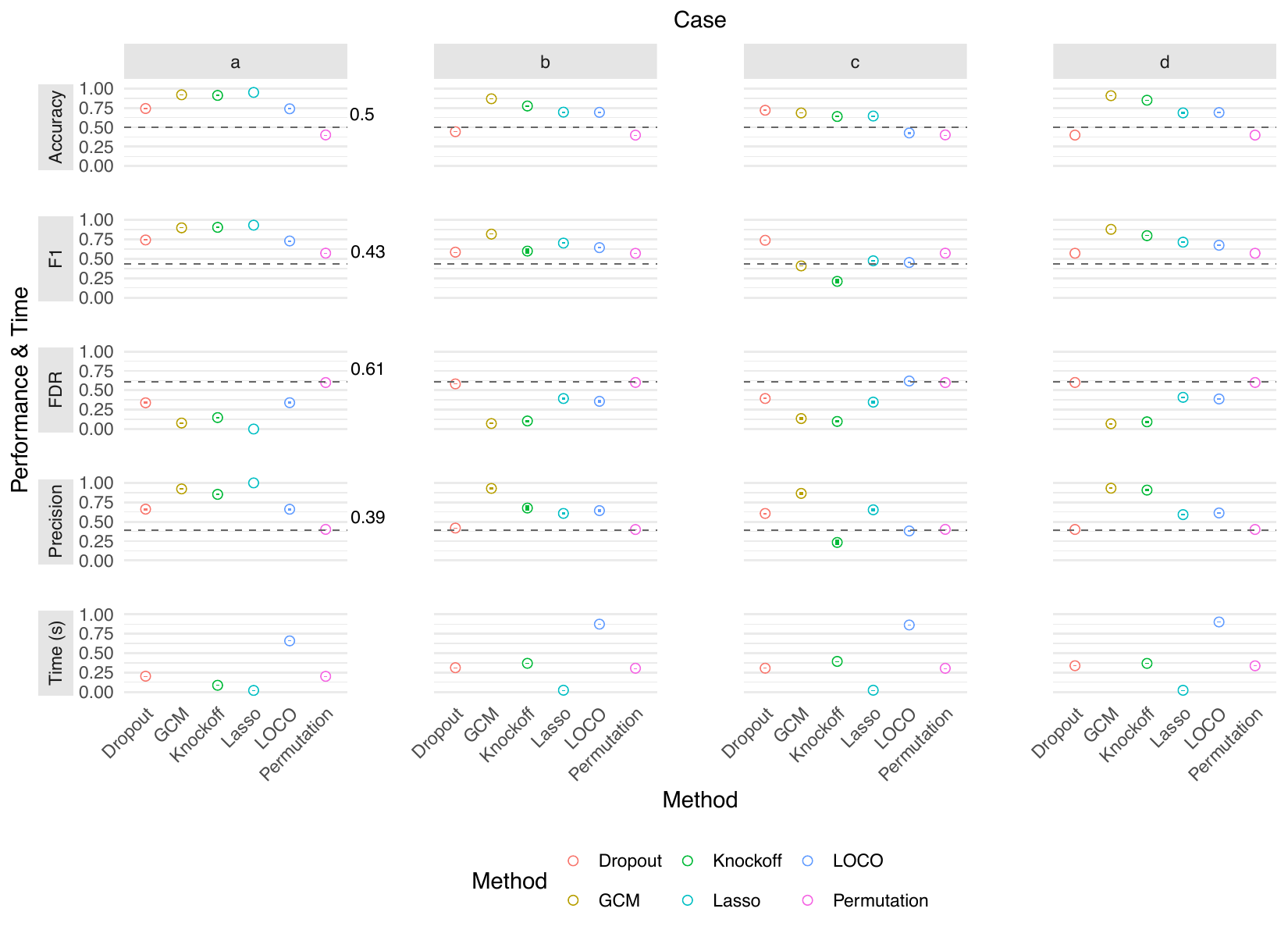}
  \caption{Averaged performance measure and time (seconds) with standard deviation across 100 simulations for lasso,  knockoff (target FDR = 0.2), GCM, LOCO, dropout and permutation univariate selection on dataset with 1000 instances using kernel SVM for (a)-(d). Dashed line: empirical random‐guessing baseline.}
  \label{performance_metric_plot_ksvm}
\end{figure}
Next, we also include another popular machine learning algorithm - kernel SVM - as a reference. For case (a) - (b) in Fig.~\ref{gcm_loco_ksvm_plots}, we use gaussian  kernel with default setting in \texttt{R} since gaussian kernel is most widely used for non-linear models with complex decision boundaries. 

We observe that GCM and LOCO show consistent results, whereas dropout and permutation methods, despite their computational efficiency, show a significant loss of type I error control in case (b), achieving high recall but low specificity. The gaussian kernel depends on all features simultaneously. Without retraining the model, any change to a feature can cause a significant difference in the learned model, making permutation and dropout methods less stable. LOCO requires approximately half the computational time of GCM, but GCM can identify more significant variables than LOCO. GCM outperforms knockoff and lasso methods in the non-linear model when the underlying model is not sparse (see Fig.~\ref{performance_metric_plot_ksvm}).
\begin{figure}[t!]
  \centering
  \includegraphics[width =\textwidth]{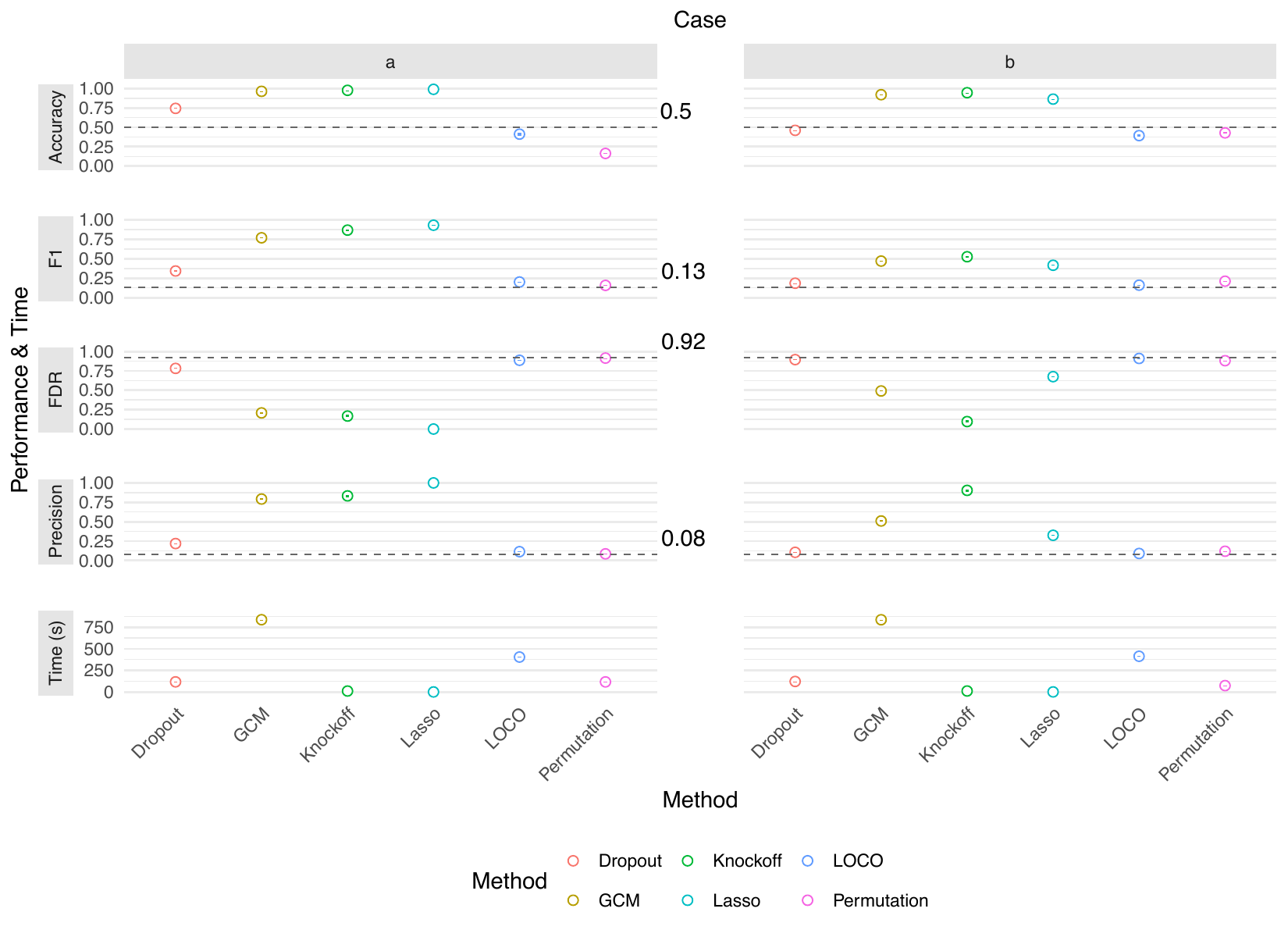}
  \caption{Averaged performance measure and time (seconds) with standard deviation across 50 simulations for lasso,  knockoff (target FDR = 0.2), GCM, LOCO, dropout and permutation univariate selection on dataset with 1000 instances and 500 features using kernel SVM for (a) and (b). Dashed line: empirical random‐guessing baseline.}
  \label{performance_metric_plot_ksvm_HD}
\end{figure}
In the high-dimensional regression setting, ksvm is implemented with the gaussian kernel. We use the default tuning parameters for setting (a), while for setting (b) we set the cost parameter to \(C=0.95\), and the setting a and b is the same as the main context Section 5.3. From Fig.~\ref{performance_metric_plot_ksvm_HD},
 Knockoff filter method and lasso work well in the sparse models, and GCM still significantly outshines other methods. Across different model training algorithms such as canonical implementations for each designated model, GBM, and kernel SVM, GCM consistently outperforms or performs as good as LOCO, indicating that GCM is a more reliable and stable approach.
\subsection{Wine quality data}
With increasing interest in wine, quality certification has become essential, relying on sensory evaluations by human experts. We aim to predict wine quality based on objective analytical tests conducted during the certification process.
The Wine Quality dataset, introduced by \cite{cortez_modeling_2009}, contains $n=1599$ samples of red wine, with $p=11$ features related to physicochemical properties (e.g., \texttt{acidity}, \texttt{alcohol}). The target variable is the wine quality score, ranging from 0 to 10, which indicates the perceived quality of the wine. 
\begin{figure}[t!]
    \centering
      \includegraphics[width=\textwidth]
    {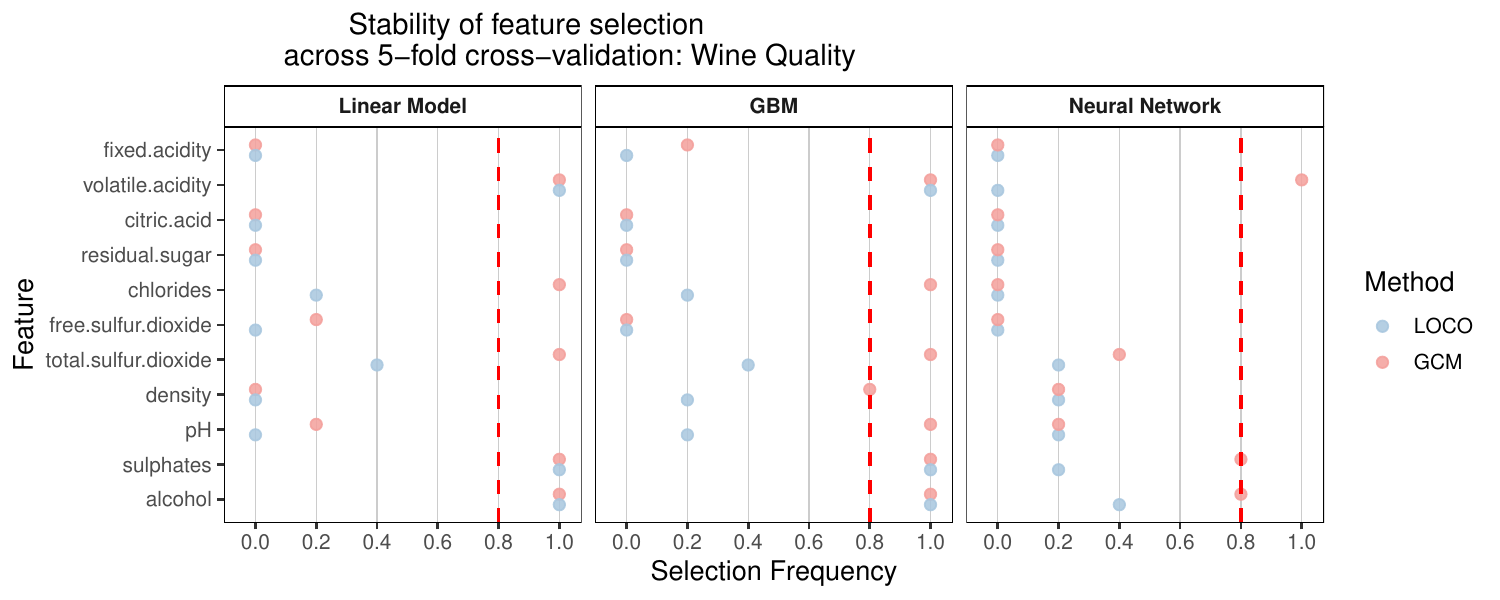}
    \caption{Selection rate of each feature selected by GCM and LOCO across 5-fold cross validation on Wine quality data, using lm, GBM and NN as the base predictor respectively. Red dashed line: targeted selection frequency.}
    \label{wine_plt}
\end{figure}
\begin{table}[t!]
    \centering
    \caption{Mean squared prediction error averaged across 5-fold cross-validation for GCM and LOCO 
on Wine quality data, using lm, GBM, NN respectively. }
\begin{tabular}{l|ccc}
\hline & \multicolumn{3}{|c}{ Models: } \\
Method & lm & GBM & NN  \\
\hline GCM &0.43 $\pm$ 0.015 & 0.41 $\pm$ 0.017 & 0.44  $\pm$ 0.026\\
LOCO &  0.44 $\pm$ 0.015 & 0.42 $\pm$ 0.017  & 0.59 $\pm$ 0.066 \\
\hline
\end{tabular}
\label{wine_tb}
\end{table}
For Wine quality data, the GBM model
uses the default setting in R. We use a fully connected neural network with  hidden layer of width $50$, trained with learning rate $5\times 10^{-4}$, weight decay $10^{-5}$, and batch size $32$. GCM constantly achieves lower MSE than LOCO across all three learners as shown in Table \ref{wine_tb}.
From Fig.~\ref{wine_plt}, we see that GCM shows more stable feature selection than LOCO. In particular, GCM consistently identifies key predictors such as alcohol, sulphates, and volatile acidity, which is the same as prior literature on wine quality analysis~\cite{longo_relating_2023}.


 \subsection{Boston housing Data}
Accurate prediction of house prices is crucial, requiring careful consideration of numerous factors that significantly influence property values. We aim to address pricing discrepancies by variable selection to predict real estate prices, providing insights for both buyers and sellers in the future. We analyze the Boston Housing dataset, originally introduced by  \cite{harrison_hedonic_1978}. This dataset contains $n=506$ observations, $p=13$ attributes of homes in Boston suburbs, along with a target variable, \texttt{medv}, representing the median value of owner-occupied homes (in thousands of dollars). 
\begin{figure}[t!]
    \centering
      \includegraphics[width=\textwidth]
    {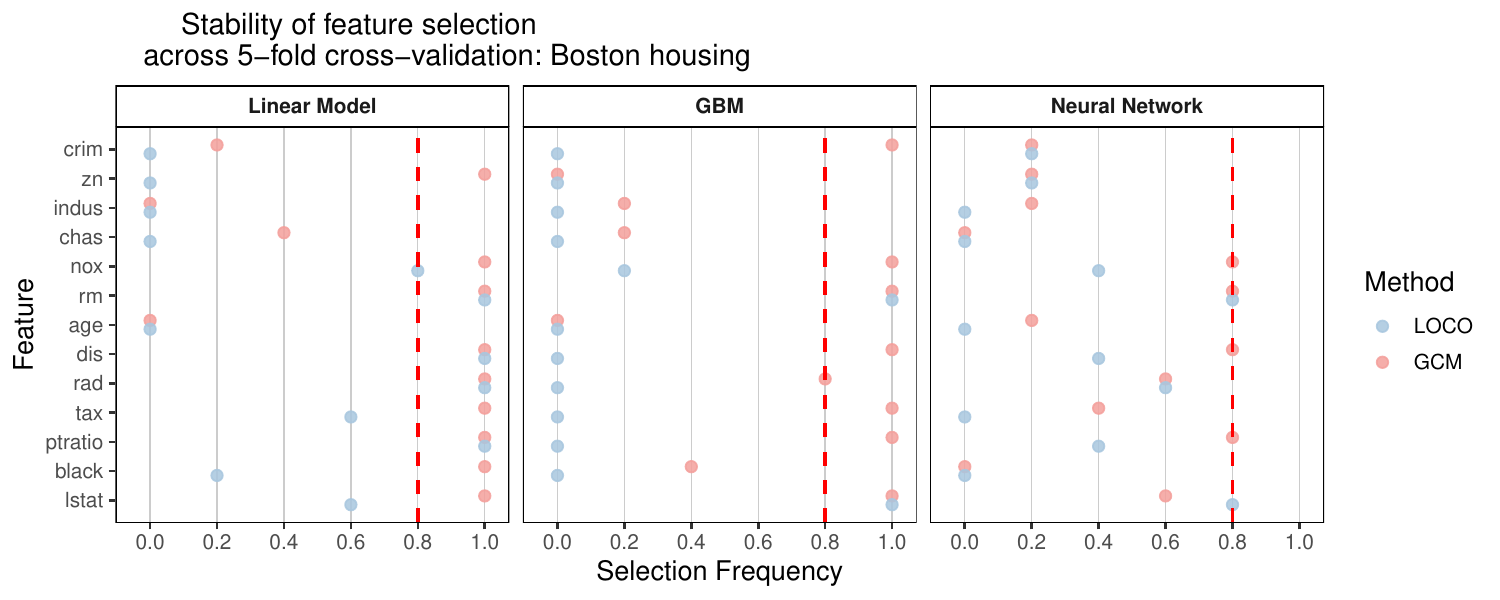}

    \caption{Selection rate of each feature selected by GCM and LOCO across 5-fold cross validation on Boston hosing data, using lm, GBM and NN as the base predictor respectively. Red dashed line: targeted selection frequency.}
    \label{boston_plt}
\end{figure}
For Boston housing data, the GBM model
uses the default setting in R. We use a fully connected neural network with hidden layer of width $32$, trained with learning rate $5\times 10^{-4}$, weight decay $10^{-3}$, and batch size $32$. From Fig.~\ref{boston_plt}, we see that the most important features selected by GCM and LOCO differ depending on the machine learning methods employed, but the mean squared prediction error of the models selected by GCM is always lower than that of LOCO (Table~\ref{boston_tb}). It can also be seen that the gradient boosted decision tree algorithm outperforms the others. Due to the robustness of gradient boosted decision tress and its ability to capture nonlinear relationships in the data, we now focus on the features selected by this model. Besides features selected by both LOCO and GCM, which indicates the significance of property details (ie: average number of rooms per house, \texttt{rm}) and the proportion of population with lower socioeconomic status (ie: \texttt{lstat}),other moderately significant features selected by GCM but not LOCO are aligned with previous studies~\cite{harrison_hedonic_1978, sanyal_boston_2022, eze_comparative_2023}.  Those features indicate the significance of proximity to major employment centers (ie: \texttt{dis}),  property tax rate (ie: \texttt{tax}), the education quality (ie: pupil-teacher ratio, \texttt{ptratio} ), etc. The results align with intuitive understanding: factors related to comfort, education, and economic level are more important when middle-income families purchase a home.  

\begin{table}[t!]
    \centering
    \caption{Mean squared prediction error averaged across 5-fold cross-validation for GCM and LOCO 
on Boston housing data, using lm, GBM, NN respectively. }
\begin{tabular}{l|ccc}
\hline & \multicolumn{3}{|c}{ Models: } \\
Method & lm & GBM & NN  \\
\hline GCM &25.79 $\pm$ 2.09&15.25 $\pm$ 1.79 & 21.21  $\pm$ 3.33\\
LOCO & 32.28 $\pm$ 4.71 & 19.17 $ \pm$ 2.37  & 23.13 $\pm$ 2.50 \\
\hline
\end{tabular}
\label{boston_tb}
\end{table}

\subsection{Crab age prediction data}
Crab age prediction is useful for commercial farming because estimating age from physical attributes helps determine the optimal harvest time and improve profit. The data provided by~\cite{gursewak_singh_sidhu_2021} is available on Kaggle, and contain 3,893 observations with \texttt{Age} as the response variable. The original predictors include \texttt{Sex}, \texttt{Length}, \texttt{Diameter}, \texttt{Height}, \texttt{Weight}, \texttt{Shucked Weight}, \texttt{Viscera Weight}, and \texttt{Shell Weight}. \texttt{Sex} is a categorical variable which has three levels: male, female, and indeterminate. We use two dummy variables to encode \texttt{Sex}, taking female as the reference level. For the Crab Age data, the GBM model uses the default setting in R. The neural network is a fully connected feed-forward model with hidden layer of 30 neurons, learning rate $5\times 10^{-4}$, weight decay $10^{-3}$, and batch size 128. The hyperparameter settings are kept consistent across LOCO and GCM. Similarly, GCM achieves lower averaged MSE than LOCO across all three 
learners as shown in Table~\ref{crab_tb}. From Fig.~\ref{crab_stability}, we see that Weight-related variables, such as \texttt{Shucked Weight} and \texttt{Shell Weight}, are selected most consistently, suggesting that body-mass measurements are significant predictors of crab age. Besides, GCM appears slightly more stable than LOCO from the selection stability perspective as it has fewer variables with intermediate selection frequencies than LOCO does.

\begin{table}[t]
    \centering
 \caption{Mean squared prediction error averaged across 5-fold cross-validation for GCM and LOCO 
on Crab Age data, using lm, GBM, NN respectively. }
\begin{tabular}{l|ccc}
\hline & \multicolumn{3}{|c}{ Models: } \\
Method & lm & GBM & NN  \\
\hline GCM &4.93 $\pm$ 0.16 & 4.73 $\pm$ 0.21 &5.82  $\pm$0.52 \\
LOCO &  5.03 $\pm$ 0.17 & 4.86$\pm$0.21  & 6.47 $\pm$ 0.57\\
\hline
\end{tabular}
\label{crab_tb}
\end{table}

\begin{figure}[t]
    \centering     \includegraphics[width=\textwidth]  {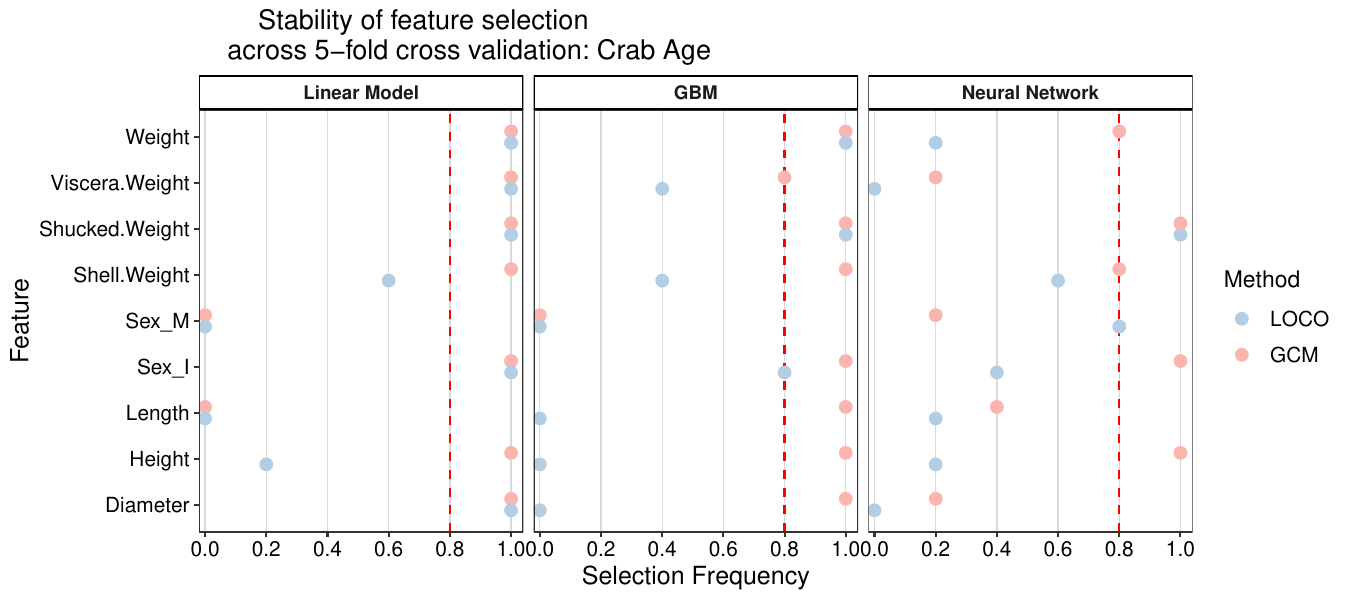}
    \caption{Selection rate of each feature selected by GCM and LOCO across 5-fold cross validation on Crab Age data, using lm, GBM and NN as the base predictor respectively. Red dashed line: targeted selection frequency.}
    \label{crab_stability}
\end{figure}

\subsection{Gas turbine $\mathrm{NO}_x$ emission data }
\begin{figure}[t!]
    \centering     \includegraphics[width=0.9\textwidth]  {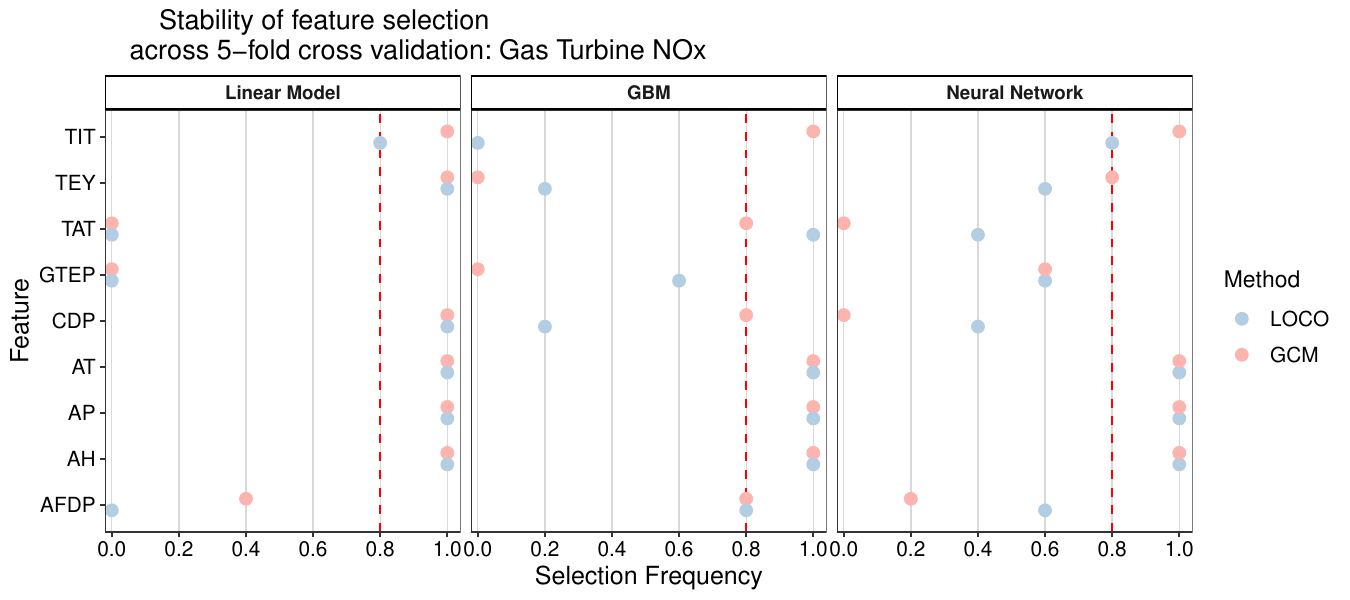}
    \caption{Selection rate of each feature selected by GCM and LOCO across 5-fold cross validation on gas turbine $\mathrm{NO}_x$ data, using lm, GBM and NN as the base predictor respectively. Red dashed line: targeted selection frequency. }
    \label{gt_stability}
\end{figure}
\begin{table}[t!]
    \centering
 \caption{Mean squared prediction error averaged across 5-fold cross-validation for GCM and LOCO 
on Gas Turbine $\mathrm{NO}_x$ data, using lm, GBM, NN respectively. }
\begin{tabular}{l|ccc}
\hline & \multicolumn{3}{|c}{ Models: } \\
Method & lm & GBM & NN  \\
\hline GCM &83.64 $\pm$ 1.92 & 39.69 $\pm$ 0.93 &38.61 $\pm$ 1.27\\
LOCO &  84.61 $\pm$ 1.73 &41.19 $\pm$ 0.73 & 38.91 $\pm$ 3.38\\
\hline
\end{tabular}
\label{gt_tb}
\end{table}
Gas turbines are significant sources of air pollution, with $\mathrm{NO}_x$ and CO as major pollutants. Accurate $\mathrm{NO}_x$ prediction can support emission monitoring and pollution control. The gas turbine dataset is publicly available in~\cite{gas_turbine_co_and_nox_emission_data}. In this study, we focus on emissions from 2013, which include 7,152 observations, and predict $\mathrm{NO}_x$ emissions concentration using nine input variables: ambient temperature, ambient pressure, ambient humidity, air filter difference pressure, gas turbine exhaust pressure, turbine inlet temperature, turbine after temperature, turbine energy yield, and compressor discharge pressure. For the Gas Turbine NO$_x$ data, the GBM model uses the default setting in R. The neural network is a fully connected feed-forward model with hidden layer of 50 neurons, learning rate $10^{-3}$, weight decay $10^{-5}$, and batch size 32. These learner settings are kept fixed across LOCO and GCM for comparison. From Table \ref{gt_tb}, we see that GCM consistently achieves the lower averaged MSE across all different machine learning learners. From Fig. \ref{gt_stability}, we see that GCM is relatively more stable than LOCO , especially for GBM and neural network learners, with fewer features having intermediate selection frequencies. Both methods frequently select \texttt{AH}, \texttt{AP}, and \texttt{AT}, while GCM tends to select additional operating variables such as \texttt{TIT}, \texttt{TEY}, and \texttt{CDP}. It is aligned with intuitive understanding since thermal NO$_x$ production is sensitive to the operating temperature, load, and pressure conditions of the turbine.

\clearpage

\end{appendices}
\end{document}